\documentclass[acmtog]{acmart}

\usepackage{booktabs} 

\setcitestyle{square}

\usepackage[ruled]{algorithm2e} 
\usepackage{overpic}
\usepackage{graphics} 

\SetAlFnt{\small}
\SetAlCapFnt{\small}
\SetAlCapNameFnt{\small}
\SetAlCapHSkip{0pt}

\setcopyright{acmlicensed}
\acmJournal{TOG}
\acmYear{2025} 
\acmVolume{44} 
\acmNumber{4} 
\acmArticle{} 
\acmMonth{8}
\acmDOI{10.1145/3730880}

\setcopyright{acmcopyright}


\graphicspath{{figures/images/}}

\usepackage{tabularx}
\usepackage{xcolor}
\usepackage{multirow}
\usepackage{hhline}
\usepackage{makecell}

\usepackage{amsthm}
\usepackage{amsmath}
\usepackage{colortbl}
\usepackage[normalem]{ulem}
\useunder{\uline}{\ul}{}
\usepackage{array}
\usepackage{booktabs}
\usepackage{tabularx} 
\newcolumntype{Y}{>{\centering\arraybackslash}X}

\usepackage{dutchcal}
\usepackage{enumitem}
\setlist[itemize]{topsep=0pt}

\definecolor{green}{rgb}{0, 0.5, 0}
\definecolor{orange}{rgb}{1.0, 0.6, 0.2}
\definecolor{red}{rgb}{1.0, 0.0, 0.0}
\definecolor{blue}{rgb}{0.0, 0.0, 1.0}
\definecolor{teal}{rgb}{0.0, 0.4, 0.4}
\definecolor{purple}{rgb}{0.65,0,0.65}
\definecolor{saffron}{rgb}{0.95,0.75,0.2}
\definecolor{turquoise}{rgb}{0.0,0.5,0.5}
\definecolor{black}{rgb}{0,0,0}

\usepackage[normalem]{ulem}

\begin{document}

\title{Designing Pin-pression Gripper and Learning its Dexterous Grasping with Online In-hand Adjustment}

\author{Hewen Xiao}
\authornote{Joint first authors.}
\affiliation{%
	\institution{Dalian University of Technology}
    \country{China}
}
\email{hewenxiao@mail.dlut.edu.cn}

\author{Xiuping Liu}
\authornotemark[1]
\affiliation{%
	\institution{Dalian University of Technology}
    \country{China}
}
\email{xpliu@dlut.edu.cn}

\author{Hang Zhao}
\authornotemark[1]
\affiliation{%
	\institution{Wuhan University}
    \country{China}
}
\email{alexfrom0815@gmail.com}

\author{Jian Liu}
\authornote{Corresponding author.}
\affiliation{%
	\institution{Shenyang University of Technology}
    \country{China}
}
\email{jianliu2006@gmail.com}

\author{Kai Xu}
\affiliation{%
	\institution{National University of Defense Technology}
    \country{China}
}
\email{kevin.kai.xu@gmail.com}

\begin{abstract}
We introduce a novel design of parallel-jaw grippers drawing inspiration from pin-pression toys. The proposed pin-pression gripper features a distinctive mechanism in which each finger integrates a 2D array of pins capable of independent extension and retraction. This unique design allows the gripper to instantaneously customize its finger’s shape to conform to the object being grasped by dynamically adjusting the extension/retraction of the pins. In addition, the gripper excels in in-hand re-orientation of objects for enhanced grasping stability again via dynamically adjusting the pins. To learn the dynamic grasping skills of pin-pression grippers, we devise a dedicated reinforcement learning algorithm with careful designs of state representation and reward shaping. To achieve a more efficient grasp-while-lift grasping mode, we propose a curriculum learning scheme. Extensive evaluations demonstrate that our design, together with the learned skills, leads to highly flexible and robust grasping with much stronger generality to unseen objects than alternatives.
We also highlight encouraging physical results of sim-to-real transfer on a physically manufactured pin-pression gripper, demonstrating the practical significance of our novel gripper design and grasping skill. 
Demonstration videos for this paper are available at \url{https://github.com/siggraph-pin-pression-gripper/pin-pression-gripper-video}.
\end{abstract}


\begin{teaserfigure}
   \begin{overpic}[width=1.0\linewidth,tics=10]{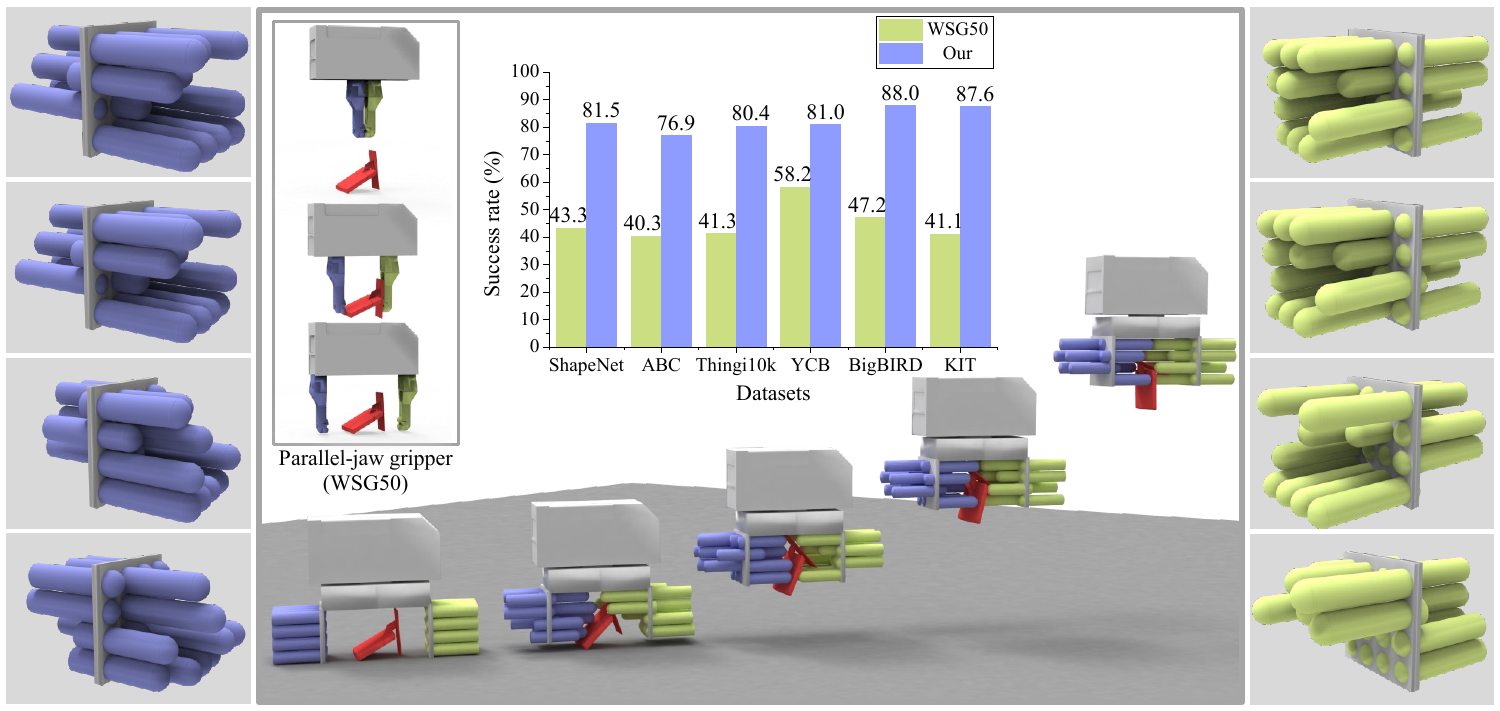}
   \end{overpic}
   \vspace{-14pt}
   \caption{
   The proposed pin-pression gripper offers object adaption and in-hand re-orientation through dynamically adjusting the extension and retraction of pins. The target object is successfully grasped and securely lifted with online pin adjustments on both fingers (shown on two sides) throughout the grasping process. As a comparison, we show the grasping of a parallel-jaw gripper (WSG50). The comparison of success rates tested on various datasets is also reported.
   }
   \label{fig:teaser}
\end{teaserfigure} 

\maketitle

\section{INTRODUCTION}\label{sec:intro}
The realm of dexterous robotic manipulation has undergone notable advancements lately. Currently, research endeavors predominantly focus on acquiring dexterous grasping skills tailored to existing gripper designs~\cite{mahler2017dex}. However, a conspicuous scarcity exists in the literature that addresses the dual challenge of simultaneous gripper designing and grasping skill learning for maximized performance. The existing works on gripper design primarily revolve around refining parallel-jaw grippers~\cite{design_overview}. 
This refinement centers on the optimization of the two fingers' shape, aiming to improve grasping performance tailored to specific object instances or categories (e.g.,~\cite{fit2form}).
Thus, the capability of these grippers can hardly be universally generalized to novel, unseen objects.
\begin{figure}[!t]
  \centering
  \begin{overpic}[width=0.8\columnwidth,tics=10]{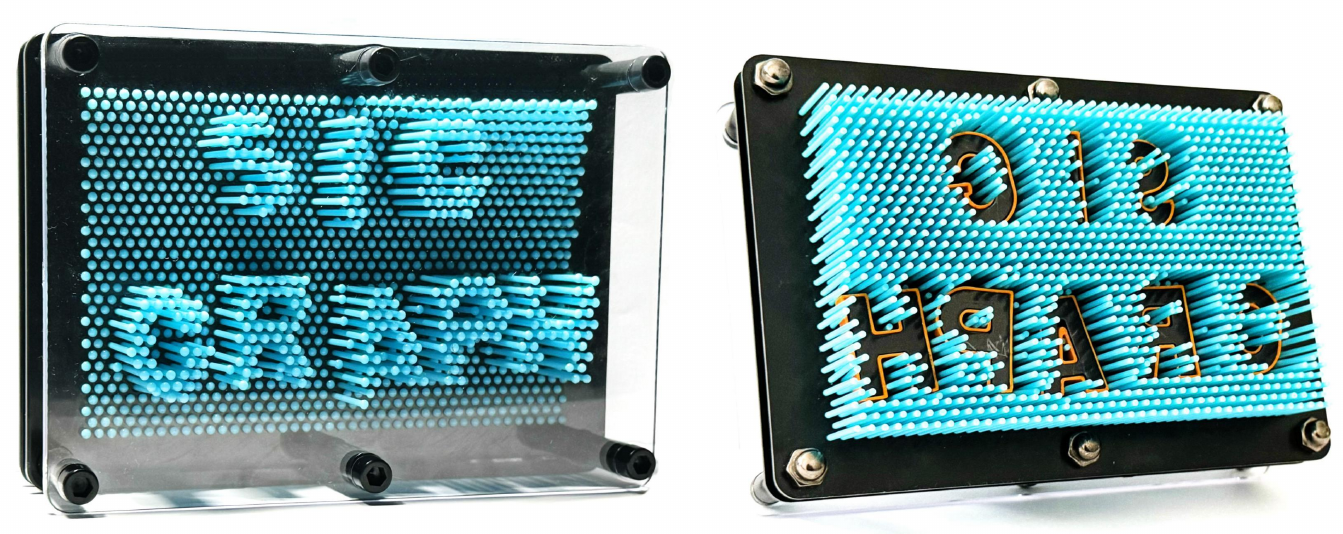}
  \end{overpic}
  \caption{A picture of a pin-pression toy. The toy features a 2D array of pins capable of independent extension and retraction.
  When pressing over the pins, an imprint of the letter is formed, which is naturally adaptive to the letter’s shape.}
   \label{fig:3D_pin_art_toy}
   \vspace{-10pt}
\end{figure}

We introduce a novel design of parallel-jaw grippers featuring dynamically adjustable finger shapes, aiming to enhance grasping adaptability to objects of arbitrary shapes. Our design draws inspiration from the captivating pin-pression toy (see Figure~\ref{fig:3D_pin_art_toy}). Our proposed gripper, coined \emph{pin-pression gripper}, features a distinctive mechanism in which each finger integrates a 2D array of pins capable of independent extension and retraction (see Figure~\ref{fig:teaser}). This design allows the gripper to instantaneously customize its fingers' shape to conform to the object being grasped by dynamically adjusting the extension and retraction of the pins. Moreover, our gripper excels in in-hand re-orientation of objects, similar to~\cite{andrychowicz2020learning}, achieving enhanced stability in grasping, again through online adjusting the pins. This design endows our gripper with both flexibility and robustness, elevating its proficiency in adeptly grasping a diverse range of objects.
A similar design dates back to~\cite{scott1985omnigripper} which, however, does not allow active adjustment of pins for target adaption. Such active adjustment requires a controller.

Learning the control skill of a pin-pression gripper for dexterous grasping and in-hand manipulation is a non-trivial task since it has $6+N$ DoFs with $N$ being the total number of pins on both fingers. In this work, we focus on a top-down grasp policy in which the gripper approaches the target object from the top, forms a closure against the object with pin adjustments, and lifts the object while performing some in-hand re-orientation to improve the grasping stability if needed. To this end, we devise a dedicated reinforcement learning (RL) procedure to learn a policy model that encompasses both grasping and in-hand manipulation and is able to generalize to unseen objects with arbitrary shapes.

We learn our grasping policy network using the off-policy RL algorithm Soft Actor-Critic (SAC)~\cite{sac}, with carefully designed state representation and rewards. In particular, the state representation of our policy model consists of the surface geometry (3D point cloud) of the target object, the interaction between the pins and the object surface (pin-surface distance and the amount and direction of protrusion of each pin), as well as the pose of the gripper. Our reward is designed to encourage successful and high-quality grasps while punishing long execution time and redundant movement of pins. 
The actions include extending and retracting pins, lifting the gripper, and stopping. The grasping policy is learned in the PyBullet simulation environment via trial and error.

There are two different modes of grasping for a pin-pression gripper: \emph{Grasp-then-Lift} (GtL) and \emph{Grasp-while-Lift} (GwL). GtL is basically a static grasping policy where the gripper forms a force closure of the object and then lifts it without further adjustment. In contrast, GwL performs pin adjustment dynamically throughout the process of grasping and lifting. While the former is easier to learn, the latter is more efficient since the gripper does not need to wait for the finish of ground adjustments. In some cases, GwL leads to more stable grasps since the gripper may find a better closure via unveiling the bottom of the target object when it is partially lifted. Albeit more efficient and effective, GwL is considerably more challenging to learn than GtL due to the more complicated grasping process. Our goal is to learn a grasping policy that could produce both GtL and GwL motions, whichever is more efficient. To do so, we opt for a curriculum learning strategy.

In particular, we adopt a two-stage training process with a properly aggregated replay buffer. In the first stage, we learn the GtL policy where the gripper is allowed only for ground adjustments, and no further adjustment is conducted once the object is lifted. After the learning, we collect all the state-action trajectories (experiences) into a self-exploration replay buffer. Bootstraped with the replay buffer, the second stage continues to learn a mixed policy admitting both GtL and GwL, while enriching the replay buffer with both experiments. To further encourage the policy to include more air adjustments, we inject the replay buffer with experiences of air-only grasping collected from a policy pre-trained with only air adjustments. Such a mixed replay buffer directs the policy model to gain proficiency in both GtL and GwL skills.

We have conducted extensive experiments to evaluate the flexibility, stability, and generality of our gripper design along with the learned grasping policy. A notable result is that our gripper attains robust and efficient grasping on unseen objects with significant variation of geometry and topology, and under arbitrary poses. We have also performed quantitative evaluations on different physics simulators such as PyBullet~\cite{coumans2019} and Isaac Gym~\cite{Makoviychuk2021IsaacGH}. We have also compared our gripper with several alternatives with state-of-the-art grasping algorithms and demonstrated consistent superiority in terms of success rate on both seen and unseen objects. Furthermore, we have showcased encouraging physical results of sim-to-real transfer through physically manufacturing a pin-pression gripper with electrically controlled pins and employing a two-stage teacher-student training paradigm. Real robot platform experiments demonstrated that our novel gripper design and grasping skill learning perform well in the real world. 
In summary, our contributions include the following.

\begin{itemize}[itemsep=0pt]
    \item A novel gripper design capable of dynamically adjusting finger shape for object shape adaption and in-hand re-orientation.
    \item A dynamic grasping policy model as well as a curriculum learning scheme achieving robust and efficient grasping with both ground and air adjustment.
    \item Extensive evaluations and physical experiments conducted on a real robot platform demonstrating the efficacy and practical significance of our gripper design and grasping skill learning.
\end{itemize}
\begin{figure}[!t]
  \centering
  \begin{overpic}[width=1.0\columnwidth,tics=10]{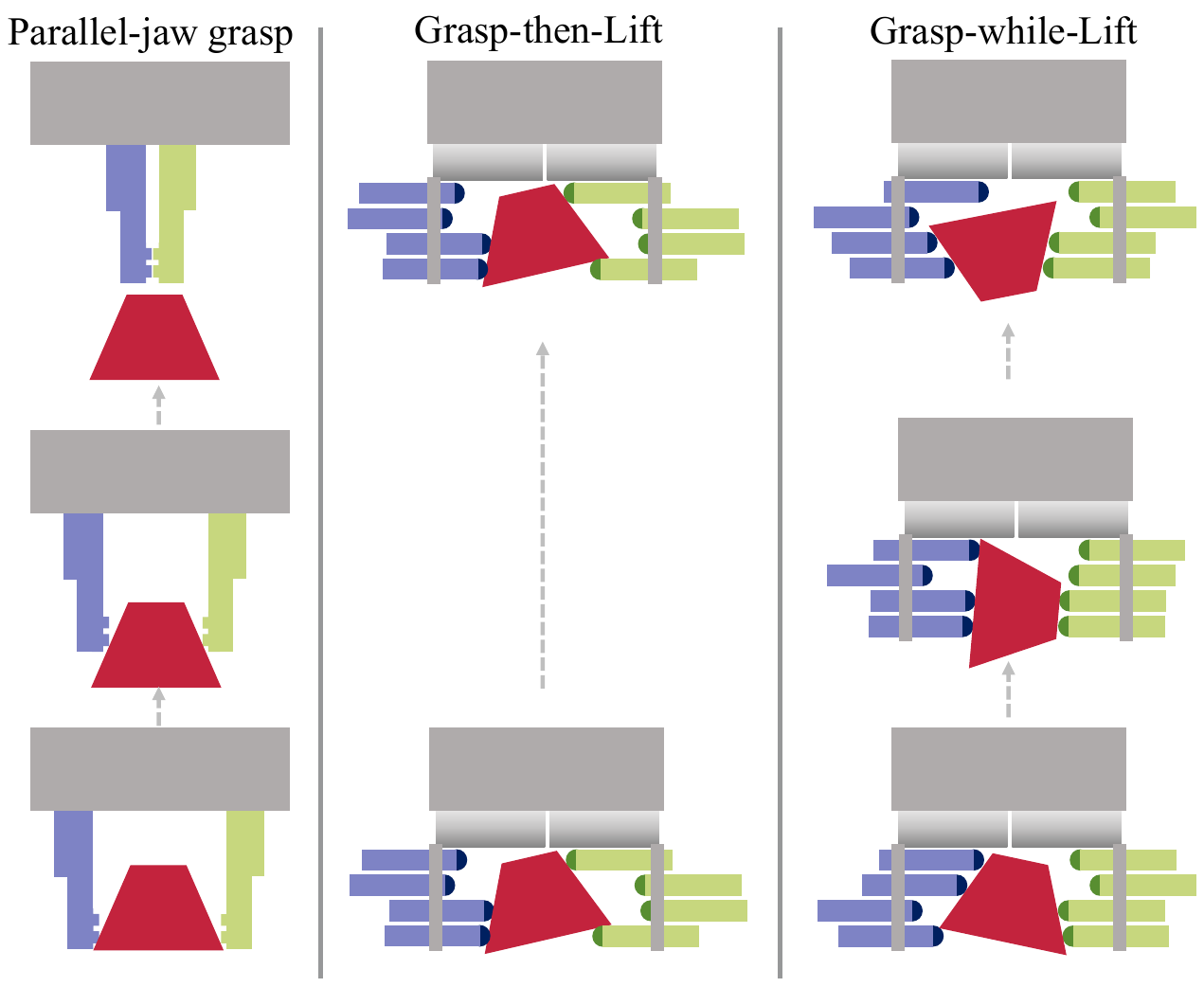}
    \put(10,-3){\small (a)}
    \put(43,-3){\small (b)}
    \put(80,-3){\small (c)}
  \end{overpic}
  \caption{The parallel-jaw gripper (a) finds difficulty in grasping the trapezoid. Our pin-pression gripper (b and c) achieves shape-adaptive grasping with dynamic adjustment of pins. In the grasp-then-lift (GtL) mode (b), the gripper can form a better closure of the trapezoid by partially lifting it. The grasp-while-lift (GwL) mode (c) allows for dynamic adjustment of pins throughout the process of grasping and lifting and therefore has the opportunity to more tightly lock the trapezoid via in-hand re-orientation.}
   \label{fig:Intro_motion}
   \vspace{-10pt}
\end{figure} 
\section{RELATED WORK}
\label{sec:related}
Finding effective ways of manipulating diverse objects has been a long-standing pursuit both in the virtual (e.g., animation) and the real world (e.g., robotics). Our work addresses a particularly challenging form of object manipulation—simultaneous grasping and reorientation—by integrating a novel gripper design with an efficient learning-based control policy for such dexterous manipulation. The unique gripper design and the associated effective policy learning together enable effective dynamic interactions between the gripper and arbitrarily-shaped objects. This highlights the synergy between designing and learning which may offer insights relevant to both robot structure design~\cite{passive_gripper, zhao2020robogrammar} and gripper-object interaction~\cite{She2022LearningHR, christen2024diffh2o}, as well as their joint optimization~\cite{geilinger2018skaterbots}.

\subsection{Robotic Gripper Design}
The existing works on gripper design primarily revolve around refining parallel-jaw grippers~\cite{design_overview},  centering on the optimization of the two fingers' shape. These parallel-jaw grippers are typically designed based on the expert experience~\cite{wsg50, Hand_e}. To enhance task performance, the evolving industrial landscape demands customized gripper shapes tailored to specific target objects. Early research focuses on modular or re-configurable design methods, selecting appropriate basic components from a finger library to assemble modular fingers or refining fingertip shapes through direct formulaic methodologies~\cite{3D_modular, Auto_Gripper, Auto_design,schroeffer2019automated, zhao2025deliberate}.
Some efforts turn to directly introducing external force 
to enable the automatic adjustment of finger shapes~\cite{robotics2019versaball, scott1985omnigripper}.
The OmniGripper~\cite{scott1985omnigripper} utilizes vertical pin arrays to wrap around and conform to the object passively, requiring a dense drive mechanism of 254 pins, and subsequent works~\cite{mo2017pin, Mo2019AUR, fu2017novel, fu2019development} have led to improvements. These gripper fingers typically operate in a fixed mode, lacking adaptability to different shapes.
 
Soft grippers also demonstrate remarkable potential in grasping tasks, leveraging the inherent flexibility and mechanical compliance of soft materials, usually through three primary mechanisms: actuation, controlled stiffness, and controlled adhesion~\cite{shintake2018soft}. The grippers like SDM Hand~\cite{dollar2010highly} address the inherent uncertainty of unstructured environments by incorporating compliance and adaptability into the hand's mechanical structure. The deformable grippers like VERSABALL~\cite{robotics2019versaball} can leverage positive and negative pressure to grasp objects. Such soft grippers do not need any feedback or computation to adjust to different shapes. However, these grippers usually impose stringent requirements on the material, such as the elastic membrane and the encapsulated granules~\cite{amend2012positive}. Our gripper does not rely on specific materials but instead achieves active shape compliance through rigid body motion.

Recently, the revolution in machine learning technology has provided new insights for task-specific gripper design, reducing reliance on expertise. Given the object to be grasped, Fit2form~\cite{fit2form} proposes a 3D generative framework that utilizes data-driven algorithms to automatically generate finger shapes for parallel-jaw grippers, enabling stable and robust grasps. Despite demonstrating satisfactory grasping performance for specific object categories, Fit2form still encounters challenges in \emph{rapidly adapting to novel objects} and requires a large amount of training data. 
ArrayBot~\cite{xue2024arraybot} utilizes $16 \times 16$  sliding pillars to vertically perceive and support the target objects for re-positioning. Unlike Arraybot's reliance on a large number of actuators, our gripper system relies solely on a $4\times 4$ array of pins to actively adjust its finger shape, thus acquiring dynamic grasping and in-hand manipulation capabilities through online pin extension and retraction. This also enables the handling of challenging objects like thin-shaped objects and trapezoids.

\subsection{Robotic Grasping Policy}

\paragraph{Analytical grasping.} 
Robotic grasping policies can be generally categorized as analytical or learning-based techniques, with the latter further divided into supervised and reinforcement learning methods.
Analytical approaches typically analyze the known geometry of a target object and obtain the optimal grasps with given quality metrics such as form closure or force closure~\cite{bicchi1995closure, zhu2003synthesis}. Some works search for grasping poses with the given configuration by minimizing torque or frictional forces~\cite{miller2004graspit} while other efforts compute the contact points or contact areas that ensure the optimal grasp quality and subsequently direct fingers to reach the optimized positions~\cite{zhu2003synthesis, Computation_curve, pan2021planning}. Contrary to the sample-based methods above, the introduction of differentiable losses~\cite{Unknown-Objects, unmodeled-objects} drives the continuous optimization techniques that apply gradient descent to directly find the optimal grasp configurations. A recent work by~\citet{Liu2020DeepDG} introduces a differentiable grasp quality metric applicable to the high-DOF gripper, enhancing the efficiency of continuously exploring optimal grasps. However, these analytical methods still face challenges in obtaining an online and real-time grasp response.

\paragraph{Supervising learning for grasping.} 
Supervised learning methods involve using deep neural networks to obtain the 2D or 3D grasp configuration for the gripper. These methods rely on the labeled data, which can be acquired through various techniques such as human demonstration~\cite{lenz2015deep, chu2018real}, physics simulation~\cite{Jacquard}, physical tests~\cite{real-robots}, and analytical computation~\cite{fang2020graspnet}. Leveraging these datasets, discriminative methods usually train a neural network to predict the grasp score of sampled grasping candidates and then select the best one~\cite{LearningHC, Dex-Net-1, mahler2017dex,song2020novel}. Generative approaches directly output a grasp configuration based on 2D or 3D information~\cite{Real-time, Closing-the-loop, mousavian20196, sharma2020deep}. Recently, to overcome the limitations of grippers' adaptability, several data-driven approaches, albeit demanding high computational costs or expensive data generation, have concentrated on learning to adapt various types of grippers and autonomously perform grasp execution~\cite{shao2020unigrasp, xu2021adagrasp,mun2023hybgrasp}. Our research enhances the adaptability of grippers by employing an innovative shape design and enables learning a dynamic grasping policy through trial and error, without the need for extensive demonstration data.

\paragraph{Reinforcement learning for grasping.}
Deep Reinforcement Learning (DRL)~\cite{drl}, offering a counterpoint to the robot grasping paradigm, has gained extensive attention in recent years. Various vision-based approaches have been developed~\cite{Vision-Based-1,du2021vision} to perceive and interact with the environment. ~\citet{Qt-opt} and ~\citet{Quantile-qt-opt} adopt the self-supervised reinforcement learning framework to perform closed-loop grasping. 
DRL also facilitates grippers' development of dynamic policies with varying mechanisms (like pick~\cite{suction-19} or push~\cite{push-22}). \citet{push-grasp-18} present a synergy parallel-jaw grasping system, which designs two fully convolutional networks that map from visual observations to pushing and grasping actions. ~\citet{push-grasp-22} employ a dual RL model to learn a policy composed of grasping and pushing. ~\citet{push-suction-19} utilize the pushing strategy to explore the environment to obtain an affordance map and adopt Deep Q-Network (DQN) to evaluate the map of suction point candidates. Recently, ~\citet{nonpre-23} design the prehensile and non-prehensile actions and employ a memory-efficient DQN to output the map of each action. Our work designs allow both grasp-then-lift and grasp-while-lift motions, enabling the dynamic and flexible grasping policy learned by the RL approach.

\paragraph{Curriculum learning.}
For complex or dexterous grasping tasks, recent studies have delved into leveraging auxiliary tasks or human demonstrations to enhance the performance of DRL~\cite{rajeswaran2017learning, zhu2018reinforcement, wang2022goal}. To solve the reach-and-grasp problem, ~\citet{She2022LearningHR} adopt the Soft Actor-Critic algorithm~\cite{sac}  to learn a grasp policy from a double replay buffer mechanism with self-exploration data and imperfect demonstrations. ~\citet{rajeswaran2017learning} explore how to combine DRL and the demonstration for learning dexterous manipulation. Integrating curriculum learning with reinforcement learning can also enrich grasping policy~\cite{ecoffet2019go, narvekar2020curriculum}. ~\citet{xu2023unidexgrasp} propose a generalizable dexterous grasping method with the PPO algorithm~\cite{ppo} and the DAgger imitation learning algorithm~\cite{dagger}, leveraging a 3-stage object curriculum learning. Further, ~\citet{UniDexGrasp++} propose a geometry-aware curriculum and iterative generalist-specialist learning method to improve performance. Similarly, ~\citet{zhang2023learning} achieve natural and robust in-hand manipulation of simple objects through deep reinforcement learning. They further facilitate the adaptation of these skills to more complex shapes through curriculum learning.
Our research focuses on tackling the challenge of gripper design and grasping skill learning concurrently. We apply curriculum learning techniques to learn the trade-off between grasp-then-lift and grasp-while-lift motion and solve the dynamic grasping problem solely based on the gripper's self-exploration experience, eliminating the need for external guidance like human demonstrations.
\section{OVERVIEW}\label{sec:overview}
\begin{figure}[!t]
  \centering
  \begin{overpic}[width=1.0\linewidth,tics=10]{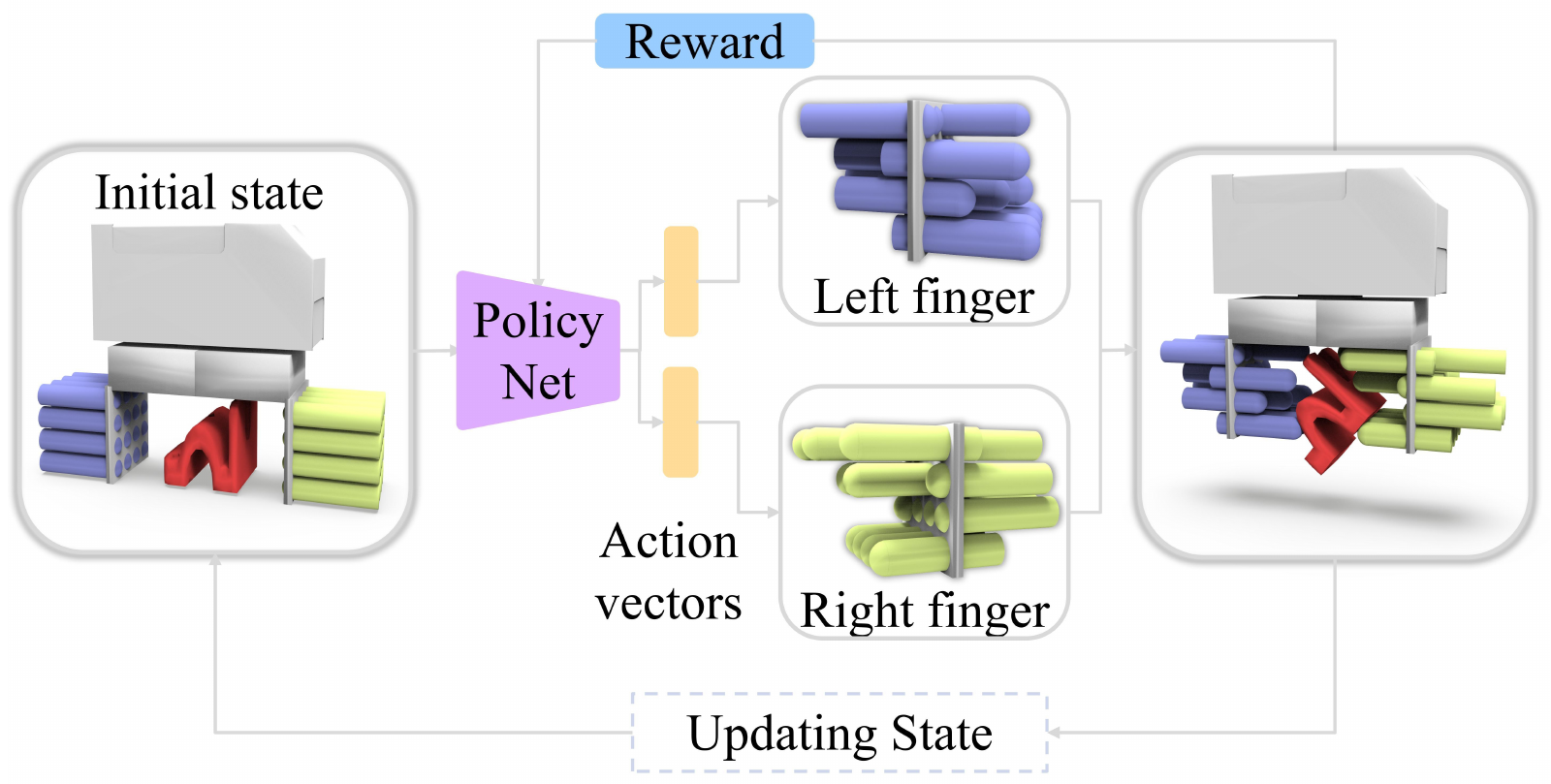}
  \end{overpic}
  \vspace{-15pt}
  \caption{Overview of our learning-based approach.  
  Our method obtains the current state information about the object, gripper, and their interaction to predict the appropriate action that moves the pins of the gripper finger in GtL or GwL grasping modes. After executing the online predicted action, the updated state is then passed through the same pipeline to predict the next action,  so that a successful grasp is gradually formed.}
   \label{fig:pipeline}
   \vspace{-10pt}
\end{figure}
\begin{figure}[!t]
  \centering
  \begin{overpic}[width=1.0\columnwidth,tics=10]{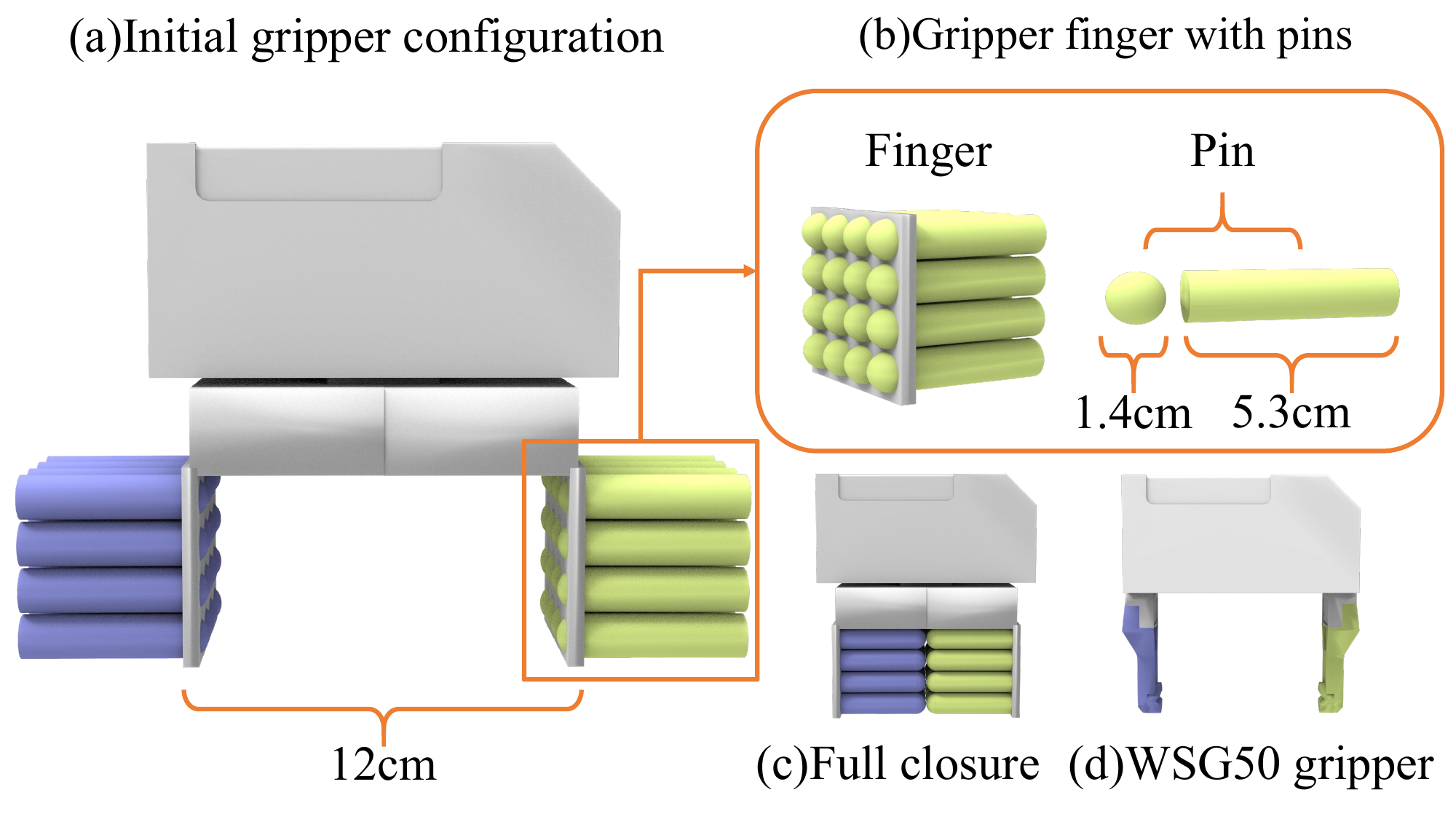}
  \end{overpic}
  \vspace{-10pt}
  \caption{The specific constitution of the parallel pin-pression gripper. The gripper finger is a 2D array of pins capable of independent extension and retraction. Each pin consists of a cylinder and a sphere.
  The gripper achieves full closure when every finger reaches its maximum forward movement.}
   \label{fig:Gripper_configuration}
   \vspace{-10pt}
\end{figure}

 
Given the target object, the goal of this work is to output a sequence of in-hand manipulations that facilitates gripper grasping with a high success rate and efficiency. We introduce a novel parallel-jaw gripper design featuring dynamic finger shape adjustments (illustrated in Figure~\ref{fig:Gripper_configuration}). Our gripper differentiates itself by a distinctive mechanism in which each finger integrates a 2D array of pins capable of independent extension and retraction.  This design allows the gripper to instantaneously customize its shape to conform to the target object by online adjustment, i.e.,  extension and retraction of the pins. We devise an effective reinforcement learning (RL) procedure to learn policy models via trial and error. A curriculum training fashion as well as the careful designs of state representation and reward shaping are adopted to enable the grasping policy to produce both \emph{grasp-then-lift} (GtL) and \emph{grasp-while-lift} (GwL) motions, possessing flexibility, stability, and generality.

The grasping task can be modeled as a Markov Decision Process (MDP)~\cite{bellman1957markovian}
$(\mathcal{S},\mathcal{A},\mathcal{P},\mathcal{R})$, where $\mathcal{S}$ represents the state space, $\mathcal{A}$ is the action space,  $\mathcal{P}:\mathcal{S} \times \mathcal{A} \mapsto \mathcal{S}$ is the state transition function, and $\mathcal{R}:\mathcal{S} \times \mathcal{A} \mapsto \mathbb{R}$ is the reward function. Figure~\ref{fig:pipeline} illustrates the pipeline of our learning-based approach. The learning starts by obtaining the current state representation $s_t$, encompassing the surface geometry of the target object, the pose of the gripper, and the interaction between the pins and the object surface.  A set of features is extracted from $s_t$ to predict the action $a_t$, adjusting the extension and retraction of the pins so that the finger shape actively adapts to the surface of the target object and ensures a physically reliable grasp. After the execution of the predicted action $a_t$, the updated state undergoes the same pipeline to determine the subsequent action. We employ a policy network learned through RL, specifically utilizing the Soft Actor-Critic method~\cite{sac}, with carefully designed grasp rewards and efficiency rewards enhancing the learning process. We also employ a two-stage curriculum learning approach, which injects the replay buffer with experiences (state-action trajectories), to achieve gripper grasping capable of both GtL and GwL motions.
\section{METHOD}\label{sec:method}
In this section, we first introduce our novel pin-pression gripper design in Section~\ref{sec:design_gripper}. Subsequently, we delve into the details of the proposed learning method, covering the state and action representations in Section~\ref{sec:state_action_representation},  the reward formulation and network architecture design in Section~\ref{sec:reward_network_design},  and the two-stage curriculum learning technique in Section~\ref{sec:curriculum}.

\subsection{Design of the Pin-pression Gripper}
\label{sec:design_gripper}
The robot end-effector is an essential interface between the robot and the target object, deeply impacting the overall performance of industrial robotic grasping tasks. The prevailing research~\cite{mahler2017dex,push-grasp-18} has predominantly focused on learning grasping policies for universal parallel-jaw grippers, where the designed shape insufficiently adapts to various target shapes. Although a recent data-driven approach~\cite{fit2form} generates the customized 3D gripper finger tailored to different target objects, it cannot adapt well to the variations in the object’s poses during the grasping process, let alone adapt to shapes it has never seen before. Consequently, the challenge of designing a gripper that adaptively attains robust and efficient online grasping on unseen objects persists.

To tackle this challenge, we strive to design an innovative gripper,  coined \emph{pin-pression gripper}, which is geometry-aware and capable of automatically adjusting its shape. Similar to Fit2form~\cite{fit2form}, we employ a general parallel-jaw gripper as the foundation since its modular design facilitates effortless customization of the shape of the gripper fingers. We observe a pin-pression toy, as depicted in Figure~\ref{fig:3D_pin_art_toy}, that inherently possesses both geometry awareness and the ability to adaptively adjust its shape. Such a toy can create impressions by pressing objects against its plastic pins, resulting in instant 3D images. Inspired by this, we integrate each gripper finger with a 2D array of pins, as illustrated in Figure~\ref{fig:Gripper_configuration}a for the initial gripper configuration.  
This grants our gripper capable of independent extension and retraction to instantaneously customize its fingers’ shape to conform to the target object.

More specifically, each gripper finger is made up of $4\times4$ cylinder pins with a height of $6$cm. The pin tip for interacting with objects is a sphere with a radius of 0.7cm. For specific configurations regarding fingers and pins, please refer to Figure~\ref{fig:Gripper_configuration}b.  
In this pin-pression gripper, the two sets of opposing pins can be extended and retracted by using a linear motion system~\cite{kurfess2005robotics}, thereby performing grasps. In practice, each pin has a maximum movement distance of $5.5$cm, and the gripper achieves full closure when every pin reaches its maximum forward movement, as depicted in Figure~\ref{fig:Gripper_configuration}c.

For the designed gripper, each grasp pose is characterized by a (6+32)-DOF configuration of the gripper. The first 6-DOF is the rigid transformation of the gripper, while the remaining 32-DOF is the stretching value, i.e., the extension length of the pins. Despite the gripper, serving as the end effector of the robotic arm, is able to move freely within a 6-DOF space,  we primarily focus on dynamic grasping with a fixed top-down grasping manner. That is, the gripper approaches the target object from the top, forms a closure against
the object through online adjustments, and lifts the object while performing in-hand re-orientation to improve the grasping stability if needed. Unlike grippers with specialized fingers, such as the WSG50 gripper depicted in Figure~\ref{fig:Gripper_configuration}d, our pin-pression gripper inherently incorporates the variances in object geometry and pose through adaptively adjusting finger shapes.

\subsection{State and Action Representation}
\label{sec:state_action_representation}

\paragraph{State representation.} 
An effective state representation is crucial for guiding the gripper to achieve a physically reliable grasp concerning the geometry and pose of the target object. It's well-known that appropriately fusing multi-modal information as input can significantly outperform single-mode data~\cite{fazeli2019see, lee2020making, miki2022learning, zhang2024multi, zhou2024multi3d}. This demonstrates that providing networks with sufficient information can enhance the performance of learned policies. Consequently, we carefully design comprehensive state representations for enhancing online grasping. We recognize that the interaction between the pins and the target object surface provides informative descriptions regarding the spatial variances. Therefore, our state vector $s_t$ at  timestep $t$ ($t$ is omitted in the subsequent discussion), involves object information $o$, gripper information $g$, and gripper-object interaction information $f$: 
\begin{equation}s = [o,g,f].\end{equation}

We elaborate on each component of the state $s$. The object information $o$ is structured as a triplet $o = [o_p,o_r,o_f]$, where $o_p$ denotes the object position, $o_r$ represents the object orientation represented using a quaternion, and $o_f$ is the feature vector derived from the object point cloud, which provides a rich representation of the object's 3D geometry characteristics~\cite{chen2023towards,su2023point,wang2024noise4denoise}. We employ DGCNN~\cite{dgcnn} to extract $o_f$, which is a 512-element vector. It can also be effortlessly adapted to use other alternative extractors owing to the modular structure of our network. In summary, the entire object information $o$ is a 519-element vector.

The gripper information $g$ for describing the gripper is formatted as a column vector, which is presented as:
\begin{equation}g = [g_p,g_p^o,g_r,g_r^o,g_{\text{stage}}, g_{\text{step}}].\end{equation}
Here $g_p$ and $g_r$ are the position and orientation of the gripper in the world frame, while $g_{p}^o$ and $g_{r}^o$  are in the object frame of reference. 
The $3$-element one-hot vector $g_{\text{stage}}$ indicates the current stage, distinguishing between ground adjustment, air adjustment, and adjustment completion. The $C$-element one-hot vector $g_{\text{step}}$ indicates the current step number. Thus, the gripper information  $g$ is a (14+3+$C$)-element vector. In practice, we set $C$ as 11.

Capturing the dynamic changes in the relationship between the gripper and the object is crucial for learning effective gripper grasping policies. Recently, \citet{She2022LearningHR} leverage Interaction Bisector Surface (IBS) to represent gripper-object interaction in the context of reach-to-grasp planning for dexterous hands. However, the computational overhead for IBS extraction is inherently substantial. Therefore, we opt for a faster and more efficient approach to characterize the gripper-object interaction, which is better suited for our pin-pression gripper.

For each pin $p$ of our gripper, we capture the following details to form the gripper-object interaction information $f$:  

\begin{itemize}
    \item Pin position $f_p \in \mathbb{R}^3$ in the world frame, represented by the tip center of $p$. 
    \item Pin position $f_p^o \in \mathbb{R}^3$ in  the object frame.
    \item Point of intersection $f_{i} \in \mathbb{R}^3$ in the world frame, obtained by extending a horizontal line from the tip to the object surface.
    \item Point of interaction $f_{i}^o \in \mathbb{R}^3$ in the object frame.
     \item Distance $f_{d} \in \mathbb{R}$ between the tip and the object surface. 
     \item Pin stretching value $f_{e} \in \mathbb{R}$.
     \item Indicator  $f_{b} \in \{-1,1\}$  informing which finger $p$ belongs to. 
    \item One-hot indicator $f_{c} \in {\{0,1\}}^{32}$ informing the pin index.
\end{itemize}

\begin{figure}[!t]
  \centering
  \begin{overpic}[width=0.8\columnwidth,tics=10]{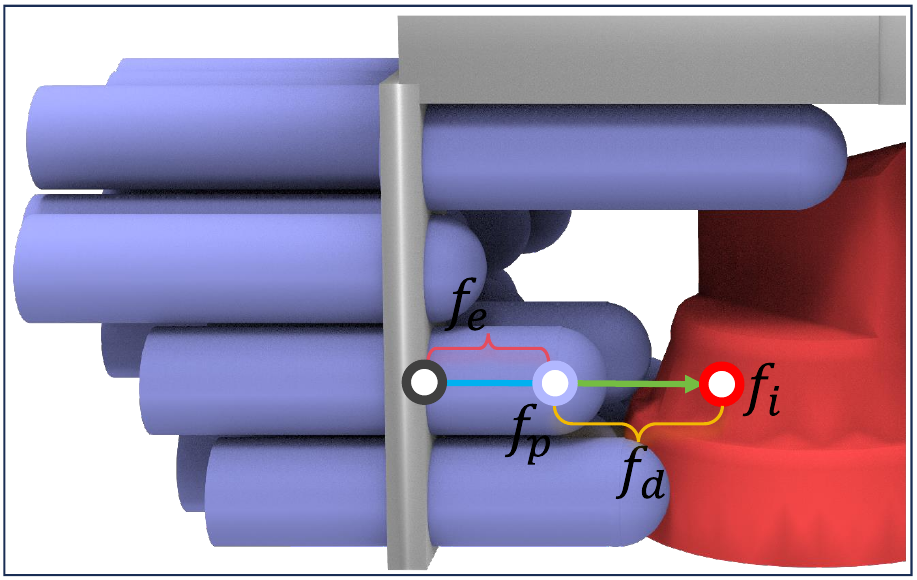}
  \end{overpic}
  \caption{Informative dynamic interaction representations of each pin $p$, including a set of relationship information between the target object and the gripper. Detailed explanations of the notations can be found in Section~\ref{sec:state_action_representation}.}
  \label{fig:state_rep}
   \vspace{-10pt}
\end{figure}

We provide visualizations of the gripper-object interaction information components in Figure~\ref{fig:state_rep}. The total length of $f$ is 1504 dimensions. The final state $s = [o,g,f]$ represents a geometry-aware depiction of the current task scene, allowing for rapid computation and accurate captures of the dynamic changes in the relationship between the gripper and the target object.

\paragraph{Action representation.}
A hybrid representation of the gripper action, integrating both GtL and GwL motions, is imperative for learning grasping skills with adaptive adjustments. To this end, the action $a$ is composed of three parts:
\begin{equation}a = [l,\chi_1,\chi_2].\end{equation}
The first part $l$ denotes the stretching value of each pin, constituting a 32-element vector confined within [0.0cm, 5.5cm]. This means that during the grasping process, our gripper can be flexibly extended and retracted for online in-hand adjustment and better grasping. The subsequent part is the switch action $\chi_1$, 
serving as a binary switch value to indicate lifting the gripper, and the potential transition from the GtL motion to the GwL motion. Here,  $\chi_1 = 0$ corresponds to the former, and  $\chi_1=1$ to the latter. The final part is the stop action $\chi_2$ serving as a termination signal for concluding the grasping. The termination criteria is elucidated in Section~\ref{sec:reward_network_design}. 
Therefore, an action is represented as a 34-element vector. 
More implementation details can be found in Section A of supplemental material.

\subsection{Reward Signal and Policy Network Design}
\label{sec:reward_network_design}
\paragraph{Reward function.}
An appropriate reward function facilitates learning adaptive grasping policies.  Given our primary purpose of ensuring the secure grasp of the target object without falling, we first formulate a \textit{grasp reward} $r_\text{grasp}$ to evaluate the grasp quality once the entire grasping procedure is completed.  We further introduce another \textit{efficiency reward} $r_\text{effici}$ to encourage minimizing unnecessary pin movements.

Similar to~\citet{She2022LearningHR}, we quantify the grasp quality by considering both the signal $G \in {\{-1,1\}}$ indicating whether the final grasp was successful, and the generalized $Q1$ analytic measure for grasp stability proposed by~\citet{Liu2020DeepDG}. Consequently, the grasp reward function is formulated as follows:
\begin{equation}r_{\text{grasp}}= \omega_{1}G +\omega_{2}Q1+r_{\text{time}},\end{equation}
where $\omega_{1} = 2000$ and $\omega_{2}=1000$ in all our experiments. An additional execution time term $r_{\text{time}}= \omega_{3}T_{\text{grasp}} +\omega_{4}T_{\text{lift}}$ is included to guide the policy in completing tasks promptly. The weight $\omega_{3}$ for grasping time  $T_{\text{grasp}}$ before lifting the object is set to -250, while the weight $\omega_{4}$ for the subsequent lifting time $T_{\text{lift}}$ is set to -50. 
Note that the entire grasp reward will only be calculated once the grasping task is finalized, i.e., $r_{\text{grasp}}$ is a terminal reward. 

To encourage efficient grasping, we would like to limit redundant pin movement as much as possible. Hence, we restrain the total length of all pin movements $r_{\text{len}}$ and the number of executed steps $r_{\text{step}}$. We formulate the efficiency reward  $r_{\text{effici}}$, serving as an intermediate reward, as follows: 
\begin{equation}
    r_{\text{effici}} =  \omega_{5}r_{\text{len}}+\omega_{6}r_{\text{step}}.
\end{equation}
In all our experiments, we consistently set $\omega_{5} = -1$ and $\omega_{6} = -10$. 

\paragraph{Network architecture and policy training.}
For the state representation, we extract features from the object point cloud and concatenate them with features obtained from gripper-object interaction and gripper information. The concatenated feature is then fed into the Multi-Layer Perceptrons (MLPs) to predict action $a$, which consists of the gripper configuration $l$,  the switch value $\chi_1$,  and the terminal value $\chi_2$. 
The grasping process terminates when either a sampled value of $\chi_2$ is greater than a threshold or a maximum number of steps is reached. We adopt DGCNN~\cite{dgcnn} for object point cloud encoding, PointNet~\cite{pointnet} for gripper-object interaction encoding, and MLPs for gripper information encoding. More details about the network architecture can be found in Section A of the supplemental material.

To train the policy network, we adopt the off-policy Soft Actor-Critic (SAC) method~\cite{sac}. Given the state $s$ as input, the actor network outputs the policy $\pi(\cdot|s, \theta)$ as a Gaussian distribution for sampling actions. The critic network takes both the state $s$ and action $a$ as inputs and outputs the state-action value for assisting the policy training. \citet{She2022LearningHR} have proved that outputting a vector by the critic network can make the training more effective than outputting a single value. Therefore, we separately approximate the accumulated grasp reward $r_{\text{grasp}}$ and efficiency reward $r_{\text{effici}}$ with a vector $(Q_{g}, Q_{e})$ output by the critic. 
The loss function for training the critic network is based on the TD update approach:
\begin{align}
L_{Q}(\beta) =  \ [&(Q_{g}(s,a;\beta)-y_{g}(r_{\text{grasp}},s',d, Q_g(s',{\widetilde{a}}',\beta')))^{2} \nonumber \\
 + & (Q_{e}(s,a;\beta)-y_{g}(r_{\text{effici}},s',d, Q_e(s',{\widetilde{a}}',\beta')))^{2} ].
\end{align}
Here the target value $y_{g}$ is formulated as:
\begin{align}
y_{g}(r,s',d, Q) = r + \ \gamma(1-d)\big[Q-\alpha\log\pi(\widetilde{a}|s';\theta)\big],
\end{align}
where $\widetilde{a}'$ is the action sampled from the policy based on the next state $s'$, $\gamma$ denotes the discount factor, and the temperature parameter \textit{$\alpha$ } represents the relative importance of the entropy term against the reward.
The training loss for the actor network is defined as:
\begin{equation}
   L_{Q}(\theta) = Q_{g}(s,\widetilde{a}(s;\theta)) + Q_{e}(s,\widetilde{a}(s;\theta)) - \alpha\log\pi(\widetilde{a}|s;\theta),
\end{equation}
where $\widetilde{a}$ represents the action sampled from the policy with state $s$ and the parameter $\theta$ will be updated.

\subsection{Curriculum Learning}
\label{sec:curriculum}
\begin{figure}[!t]
  \centering
  \begin{overpic}[width=1.0\columnwidth,tics=10]{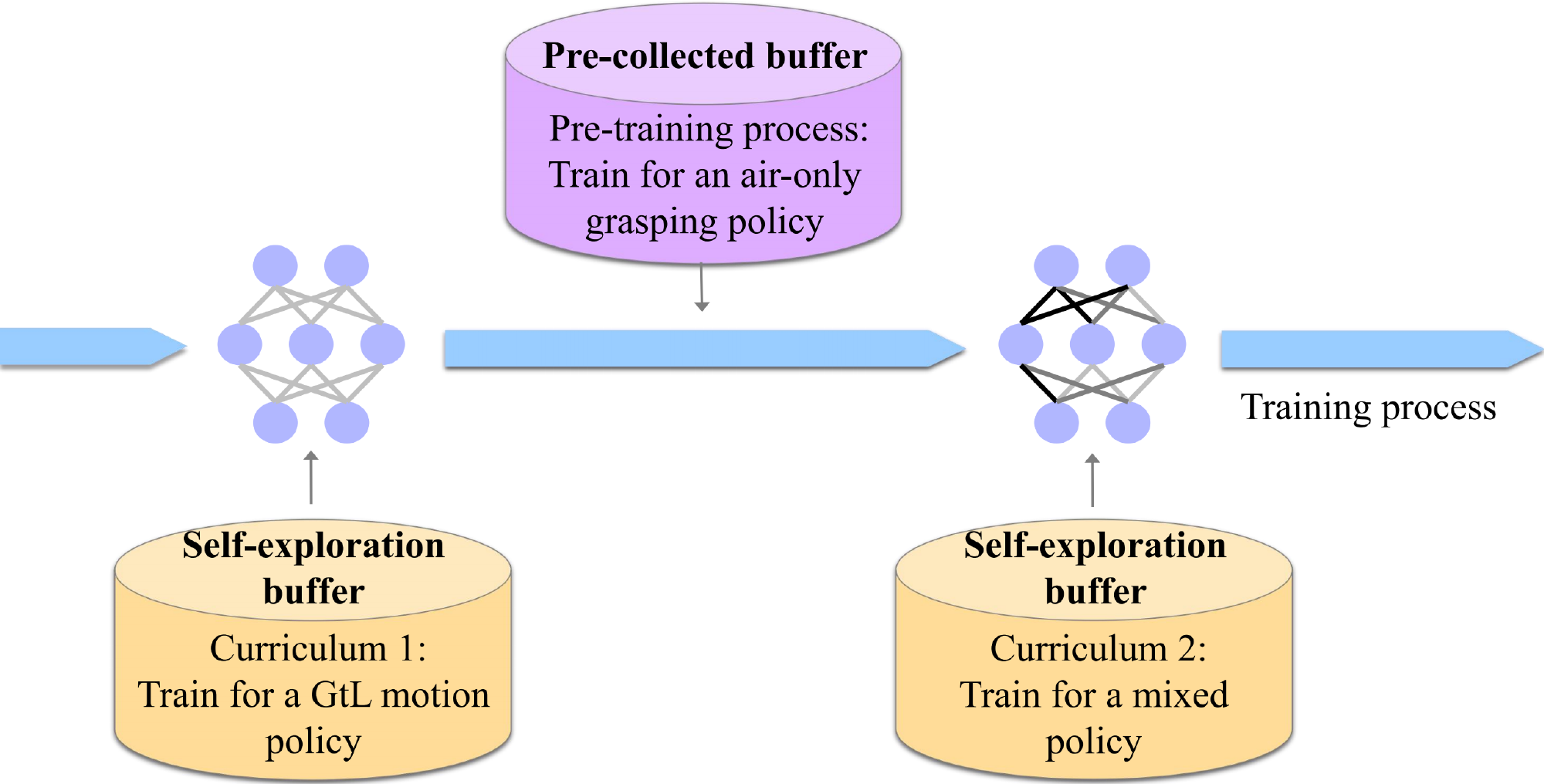}
  \end{overpic}
  \caption{Our learning framework consists of a two-stage curriculum (colored in yellow) and a pre-training process (in purple) for collecting GwL experiences. The curriculum learning progressively guides the policy that extends the ability of gripper manipulation from GtL motion to GwL motion. 
  To further encourage the policy to include more air adjustments, we inject the replay buffer with experiences of air-only grasping, collected from a policy pre-trained with only air adjustments in the learning of the second stage.
  }
   \label{fig:Method_CL_1}
   \vspace{-10pt}
\end{figure} 
In order to allow our pin-pression gripper to adaptively grasp objects with both GtL and GwL modes, we adopt a two-stage curriculum learning method with a properly aggregated replay buffer. The curriculum learning gradually bootstraps the policy to gain grasping proficiency, progressing the grasp from the GtL motion to the GwL motion.
We illustrate our learning process in Figure~\ref{fig:Method_CL_1}. In the first stage of the curriculum, we learn the GtL policy where the gripper is allowed only for ground adjustments, and no further adjustment is allowed once the object is lifted. All experiences in this stage are recorded into a self-exploration replay buffer, which also assists the second stage curriculum. The second stage continues to learn a mixed policy admitting both GtL and GwL modes while enriching the replay buffer with both experiments. In addition to the GtL data inherited from the first stage, we also inject the replay buffer with experiences of air-only grasping collected from a policy pre-trained with only air adjustments (colored purple in Figure~\ref{fig:Method_CL_1}) to further encourage the policy to include more air adjustments. Such a mixed replay buffer directs the policy model to gain proficiency in both GtL and GwL skills.

More specifically, we utilize two replay buffers for each curriculum learning stage: one for the self-exploration data with a maximal size set at $m=50000$, and the other for the pre-collected  GwL data with a maximal size set at $n=5000$. Each replay buffer is maintained as a first-in, first-out queue. We first initialize an empty buffer to start the first stage curriculum and continuously append experiences to the self-exploration buffer. In the second stage of learning, we leverage both the self-exploration buffer and the pre-collected buffer to enhance the policy's capability for cross-motion grasping. Throughout the training phase, the self-exploration buffer continuously updates with new experiences, while the pre-collected buffer remains fixed without further updates. The probability of sampling data from each buffer is directly proportional to their capacity. Our two-stage training approach enables the network to capitalize on abundant experiences and strike a balance between success rate and task execution cost.

\section{Experiment}\label{sec:experimental}
We first introduce our experimental settings in Section~\ref{sec:5_1}. Our method achieves notable performance in online grasping, rooted in the careful designs of action planning, state representations, and reward shaping. We conduct ablation studies in Section~\ref{sec:5_2} to investigate the contributions of these components. Subsequently, we provide comparisons with alternative methods in Section~\ref{sec:5_3} to elucidate the efficacy of our approach, particularly when compared with the passive grasping method. Finally, in Section~\ref{sec:5_4}, we delve into a more in-depth analysis of our pin-pression gripper and examine the performance of both sim-to-sim transfer and sim-to-real transfer under a different physics engine and on a real robot platform, respectively.

\subsection{Experimental Settings}
\label{sec:5_1}
\paragraph{Data preparation.}
To train a reliable policy for dynamically grasping 3D shapes of variations in geometry and topology, it’s necessary to generate a target object dataset containing diverse shapes with appropriate sizes while compatible with the gripper. Specifically, we collect our target objects from the  3DNet dataset~\cite{6225116}, the BigBIRD dataset~\cite{bigbird}, the ShapeNet dataset~\cite{shapenet}, the YCB dataset~\cite{ycb}, the GD dataset~\cite{7139793},  the KIT dataset~\cite{kit}, and the Thingi10K dataset~\cite{think10k}. These datasets contain abundant complex shapes from household objects to industrial components. For each target object, we settle it in a random orientation on the ground. We subsequently resize it to fit within a bounding box with a maximum side length of $5.5$cm, centered at the initial grasp location. The resized object is then incorporated into our object dataset used for grasping. The resulting object dataset contains a total of $543$ objects, with $446$ allocated for training and $97$ for testing. For a thorough gallery of object dataset, please see Figure 12 in supplemental material. We further test our method on over $5000$ unseen objects collected from various datasets to demonstrate the generalization superiority of our pin-pression gripper compared to other alternatives.
We also collect and contribute a set of $108$  objects challenging for passively grasping approaches, categorized as the Chal-H and Chal-T datasets, to emphasize the necessity of the adaptive grasping approach.
\begin{table*}[ht]
\centering
\caption{Ablation studies of our method, where $S$ denotes the success rate of the entire test or train set; $Q1$ is the mean generalized $Q1$ value of all successful grasps; $T$ presents the average running time for all objects, totally 97 objects for the testing set and  446  for the  training set; GwL (\%) is the ratio of the testing objects using GwL mode; GtL (\%) means the ratio of the grasping objects using GtL mode; 
We bold the best results in terms of $S$, $Q1$, and $T$.}
\begin{tabularx}{\textwidth}{X|
 >{\centering\arraybackslash}X|
 >{\centering\arraybackslash}X|
 >{\centering\arraybackslash}X|
 >{\centering\arraybackslash}X|
 >{\centering\arraybackslash}X|
 >{\centering\arraybackslash}X|
 >{\centering\arraybackslash}X|
 >{\centering\arraybackslash}X|
 >{\centering\arraybackslash}X|
 >{\centering\arraybackslash}X}
\hline
& \multicolumn{5}{c|}{Testing set (Unseen objects)}
& \multicolumn{5}{c}{Training set (Seen objects)}
\\ 
\hline
Method
& $S (\%) \uparrow$ 
& $Q1 \uparrow$ 
& $T  (s) \downarrow$
& GwL $(\%)$ 
& GtL $(\%)$ 
& $S (\%) \uparrow$ 
& $Q1 \uparrow$ 
& $T (s) \downarrow$  
& GwL $(\%)$ 
& GtL $(\%)$ 
\\ 

\hline 
\hline
Ours
& \multicolumn{1}{c|}{82.47}  
& \multicolumn{1}{c|}{0.3456}  
& \multicolumn{1}{c|}{6.54} 
& \multicolumn{1}{c|}{64.95}  
& \multicolumn{1}{c|}{35.05}  
& \multicolumn{1}{c|}{82.96} 
& \multicolumn{1}{c|}{0.3506}   
& \multicolumn{1}{c|}{6.30} 
& \multicolumn{1}{c|}{71.30}   
& \multicolumn{1}{c}{28.70}
\\ 
\hline
\multicolumn{1}{l|}{Ours ($\chi_1=0$)} 
& \multicolumn{1}{c|}{\textbf{84.53}}   
& \multicolumn{1}{c|}{0.3587}  
& \multicolumn{1}{c|}{7.65} 
& \multicolumn{1}{c|}{0.00}      
& \multicolumn{1}{c|}{100.00}
& \multicolumn{1}{c|}{\textbf{88.56}} 
& \multicolumn{1}{c|}{0.3561}      
& \multicolumn{1}{c|}{7.59}  
& \multicolumn{1}{c|}{0.00}      
& \multicolumn{1}{c}{100.00}  
\\ 
\hline
\multicolumn{1}{l|}{Ours ($\chi_1=1$)}
& \multicolumn{1}{c|}{61.86}   
& \multicolumn{1}{c|}{0.3603}  
& \multicolumn{1}{c|}{\textbf{6.00}} 
& \multicolumn{1}{c|}{100.00}    
& \multicolumn{1}{c|}{0.00}     
& \multicolumn{1}{c|}{74.44} 
& \multicolumn{1}{c|}{\textbf{0.3586}}      
& \multicolumn{1}{c|}{\textbf{6.00}} 
& \multicolumn{1}{c|}{100.00}    
& \multicolumn{1}{c}{0.00}    
\\ 
\hline\hline
\multicolumn{1}{l|}{w/o $G$ 
}   
& \multicolumn{1}{c|}{58.76}   
& \multicolumn{1}{c|}{0.3495}  
& \multicolumn{1}{c|}{6.01}     
& \multicolumn{1}{c|}{98.97}  
& \multicolumn{1}{c|}{1.03}
& \multicolumn{1}{c|}{52.24} 
& \multicolumn{1}{c|}{0.3515}      
& \multicolumn{1}{c|}{6.01} 
& \multicolumn{1}{c|}{99.77}  
& \multicolumn{1}{c}{0.23}  
\\ 
\hline
\multicolumn{1}{l|}{w/o $Q1$ 
}        
& \multicolumn{1}{c|}{78.35}   
& \multicolumn{1}{c|}{0.3412}  
& \multicolumn{1}{c|}{6.53}     
& \multicolumn{1}{c|}{78.35}  
& \multicolumn{1}{c|}{21.65}  
& \multicolumn{1}{c|}{81.16}      
& \multicolumn{1}{c|}{0.3503}      
& \multicolumn{1}{c|}{6.39}          
& \multicolumn{1}{c|}{70.63}       
& \multicolumn{1}{c}{29.37}      
\\ 
\hline
\multicolumn{1}{l|}{w/o $r_{\text{time}}$}  
& \multicolumn{1}{c|}{78.35}   
& \multicolumn{1}{c|}{0.3569}  
& \multicolumn{1}{c|}{9.80}     
& \multicolumn{1}{c|}{30.93}  
& \multicolumn{1}{c|}{69.07} 
& \multicolumn{1}{c|}{80.72} 
& \multicolumn{1}{c|}{0.3474}      
& \multicolumn{1}{c|}{9.07}  
& \multicolumn{1}{c|}{44.84}  
& \multicolumn{1}{c}{55.16} 
\\ 
\hline
\multicolumn{1}{l|}{w/o $r_{\text{len}}$}  
& \multicolumn{1}{c|}{78.35}   
& \multicolumn{1}{c|}{0.3554}  
& \multicolumn{1}{c|}{6.70}  
& \multicolumn{1}{c|}{50.52}  
& \multicolumn{1}{c|}{49.48}  
& \multicolumn{1}{c|}{80.49} 
& \multicolumn{1}{c|}{0.3511}      
& \multicolumn{1}{c|}{6.48} 
& \multicolumn{1}{c|}{58.97}       
& \multicolumn{1}{c}{41.03}
\\ 
\hline
\multicolumn{1}{l|}{w/o $r_{\text{step}}$}  
& \multicolumn{1}{c|}{76.29}   
& \multicolumn{1}{c|}{0.3545}  
& \multicolumn{1}{c|}{6.75}  
& \multicolumn{1}{c|}{11.34}  
& \multicolumn{1}{c|}{88.66}
& \multicolumn{1}{c|}{72.42} 
& \multicolumn{1}{c|}{0.3483}      
& \multicolumn{1}{c|}{6.72} 
& \multicolumn{1}{c|}{21.75}      
& \multicolumn{1}{c}{78.25}   
\\ 
\hline\hline 
\multicolumn{1}{l|}{w/o $o$}         
& \multicolumn{1}{c|}{72.16}       
& \multicolumn{1}{c|}{0.3385}  
& \multicolumn{1}{c|}{6.21}          
& \multicolumn{1}{c|}{75.26}       
& \multicolumn{1}{c|}{24.74}     
& \multicolumn{1}{c|}{78.03}       
& \multicolumn{1}{c|}{0.3524}      
& \multicolumn{1}{c|}{6.23}           
& \multicolumn{1}{c|}{69.73}        
& \multicolumn{1}{c}{30.27}       
\\ 
\hline
\multicolumn{1}{l|}{w/o $g$}         
& \multicolumn{1}{c|}{81.44}       
& \multicolumn{1}{c|}{0.3471}  
& \multicolumn{1}{c|}{7.55}          
& \multicolumn{1}{c|}{0.00}        
& \multicolumn{1}{c|}{100.00}        
& \multicolumn{1}{c|}{74.89}       
& \multicolumn{1}{c|}{0.3446}      
& \multicolumn{1}{c|}{6.81}           
& \multicolumn{1}{c|}{0.00}        
& \multicolumn{1}{c}{100.00}       
\\ 
\hline
\multicolumn{1}{l|}{w/o $f$}         
& \multicolumn{1}{c|}{75.25}   
& \multicolumn{1}{c|}{0.3550}  
& \multicolumn{1}{c|}{6.71}     
& \multicolumn{1}{c|}{5.15}   
& \multicolumn{1}{c|}{94.85} 
& \multicolumn{1}{c|}{79.37} 
& \multicolumn{1}{c|}{0.3512} 
& \multicolumn{1}{c|}{6.67}     
& \multicolumn{1}{c|}{20.63}  
& \multicolumn{1}{c}{79.37}
\\
\hline
\multicolumn{1}{l|}{only $f$}        
& \multicolumn{1}{c|}{69.07} 
& \multicolumn{1}{c|}{0.3602}  
& \multicolumn{1}{c|}{6.45}  
& \multicolumn{1}{c|}{57.73}  
& \multicolumn{1}{c|}{42.27}  
& \multicolumn{1}{c|}{78.48}       
& \multicolumn{1}{c|}{0.3533}      
& \multicolumn{1}{c|}{6.55}           
& \multicolumn{1}{c|}{59.42}        
& \multicolumn{1}{c}{40.58}       
\\ 
\hline
\multicolumn{1}{l|}{w/o $world$}         
& \multicolumn{1}{c|}{73.20}   
& \multicolumn{1}{c|}{0.3384}  
& \multicolumn{1}{c|}{6.23}     
& \multicolumn{1}{c|}{89.69}   
& \multicolumn{1}{c|}{10.31} 
& \multicolumn{1}{c|}{79.82} 
& \multicolumn{1}{c|}{0.3534} 
& \multicolumn{1}{c|}{6.28}     
& \multicolumn{1}{c|}{79.37}  
& \multicolumn{1}{c}{20.63}
\\ 
\hline
\multicolumn{1}{l|}{w/o $one\ hot$}         
& \multicolumn{1}{c|}{79.38}   
& \multicolumn{1}{c|}{0.3563}  
& \multicolumn{1}{c|}{6.80}     
& \multicolumn{1}{c|}{87.63}   
& \multicolumn{1}{c|}{12.37} 
& \multicolumn{1}{c|}{77.80} 
& \multicolumn{1}{c|}{0.3489} 
& \multicolumn{1}{c|}{6.72}     
& \multicolumn{1}{c|}{80.49}  
& \multicolumn{1}{c}{19.51}
\\ 
\hline
\multicolumn{1}{l|}{w/o $pin\ pos$}         
& \multicolumn{1}{c|}{77.32}   
& \multicolumn{1}{c|}{0.3550}  
& \multicolumn{1}{c|}{7.36}     
& \multicolumn{1}{c|}{18.56}   
& \multicolumn{1}{c|}{81.44} 
& \multicolumn{1}{c|}{75.11} 
& \multicolumn{1}{c|}{0.3542} 
& \multicolumn{1}{c|}{7.03}     
& \multicolumn{1}{c|}{27.13}  
& \multicolumn{1}{c}{72.87}    
\\ 
\hline
\multicolumn{1}{l|}{RGB-D image} 
& \multicolumn{1}{c|}{73.19}  
& \multicolumn{1}{c|}{\textbf{0.3609}}  
& \multicolumn{1}{c|}{6.87}    
& \multicolumn{1}{c|}{15.46}   
& \multicolumn{1}{c|}{84.54}
& \multicolumn{1}{c|}{71.75} 
& \multicolumn{1}{c|}{0.3578} 
& \multicolumn{1}{c|}{6.79}   
& \multicolumn{1}{c|}{24.22}  
& \multicolumn{1}{c}{75.78}  
\\
\hline\hline 
\multicolumn{1}{l|}{w/o $CL$}       
& \multicolumn{1}{c|}{70.10}       
& \multicolumn{1}{c|}{0.3537}  
& \multicolumn{1}{c|}{6.06}          
& \multicolumn{1}{c|}{95.88}        
& \multicolumn{1}{c|}{4.12}        
& \multicolumn{1}{c|}{74.66}       
& \multicolumn{1}{c|}{0.3567}      
& \multicolumn{1}{c|}{6.07}          
& \multicolumn{1}{c|}{93.72}        
& \multicolumn{1}{c}{6.28}         
\\ 
\hline
\multicolumn{1}{l|}{w/o $pre$}  
& \multicolumn{1}{c|}{80.41} 
& \multicolumn{1}{c|}{0.3375}  
& \multicolumn{1}{c|}{6.76} 
& \multicolumn{1}{c|}{57.73}  
& \multicolumn{1}{c|}{42.27}
& \multicolumn{1}{c|}{81.84} 
& \multicolumn{1}{c|}{0.3567}      
& \multicolumn{1}{c|}{6.61}  
& \multicolumn{1}{c|}{59.42}        
& \multicolumn{1}{c}{40.58}     
\\ 
\hline
\multicolumn{1}{l|}{w/o $CL\  \& \ pre$}      
& \multicolumn{1}{c|}{71.13} 
& \multicolumn{1}{c|}{0.3457}  
& \multicolumn{1}{c|}{6.19}  
& \multicolumn{1}{c|}{80.41}  
& \multicolumn{1}{c|}{19.59} 
& \multicolumn{1}{c|}{78.48} 
& \multicolumn{1}{c|}{0.3531}      
& \multicolumn{1}{c|}{6.20}  
& \multicolumn{1}{c|}{79.15}        
& \multicolumn{1}{c}{20.85}                  
\\ 
\hline
\multicolumn{1}{l|}{Reward $CL$}   
& \multicolumn{1}{c|}{78.35} 
& \multicolumn{1}{c|}{0.3345}  
& \multicolumn{1}{c|}{7.14} 
& \multicolumn{1}{c|}{21.65}  
& \multicolumn{1}{c|}{78.35} 
& \multicolumn{1}{c|}{79.82} 
& \multicolumn{1}{c|}{0.3401}      
& \multicolumn{1}{c|}{7.13}  
& \multicolumn{1}{c|}{21.52}        
& \multicolumn{1}{c}{78.48}
\\
\hline
\end{tabularx}
\label{tab:ablation_table}
\vspace{-7pt}
\end{table*}

\paragraph{Simulation setup.}
We evaluate the grasping performance in a simulation environment established using the PyBullet physics simulator~\cite{coumans2019}. Within this environment, a flat surface serves as the ground, an object is sampled from the target object dataset, and our pin-pression gripper is present. Before executing the simulation, the gripper is translated to a fixed position with its initial configuration, and the object is placed between the two fingers of the gripper. Considering that variations in end load can impact the grasping outcome, we assume a consistent mass of all target objects to $50$ grams each with a uniform density to focus on the gripper's adaptability to the target object geometry. Following  Fit2form~\cite{fit2form}, we set the lateral friction coefficient of $0.2$, and the rolling friction coefficient of $0.001$. We simulate the grasping process by manipulating the pins with a force of $0.5$N applied during grasping. Our simulation environment simultaneously allows both the GtL and GwL grasping modes. To verify grasp success, we lift the object to a height of 30cm and then check if it has dropped. After a simulation episode is done, the gripper is reset to its initial state.

\paragraph{Evaluation metrics.}
As discussed in Section~\ref{sec:method}, our objective is to achieve adaptive grasping with both effectiveness and efficiency. To evaluate the grasp quality, we compute the \emph{grasp success rate (S)} for the entire testing set, with the simulator determining whether the target object is dropped or not after the grasping task concludes. For all successful grasps, we calculate the average \emph{generalized Q1}~\cite{Liu2020DeepDG} as an analytical metric representing the overall physical stability. The $Q1$ value is greater than or equal to 0, and a higher value indicates a more stable grasp. To measure the grasping efficiency, we record \emph{the running time (T)}. We also measure the ratios of grasps through the GtL and GwL modes to intuitively understand the gripper's grasping behavior. 

\subsection{Ablations on RL Components}
\label{sec:5_2}
To verify the impact of each component, we conduct ablation studies by systematically removing several key components. The corresponding quantitative results are presented in Table~\ref{tab:ablation_table}. 
We use the full version of our method as a benchmark, denoted as ‘Ours’. 

\paragraph{Flexibility of action design.}
Implementing an action design that incorporates both GtL and GwL motions is essential for enabling the policy network to learn adaptive grasping skills that are both effective and efficient. We individually train the policies under each grasping mode. As depicted in Table~\ref{tab:ablation_table}, when only GtL motion is allowed for execution,  represented as ‘Ours ($\chi_1=0$)’, the policy network achieves the highest success rate of $84.53\%$ on the testing set but exhibits the lowest efficiency in terms of grasping task running time. This is primarily attributed to decoupling the actions of grasping and lifting objects into two independent stages, leading to additional task running costs about waiting for the finish of ground adjustments. On the other hand, when only GwL motion is considered, denoted as ‘Ours ($\chi_1=1$)’, the policy network yields the lowest success rate of $61.86\%$ on the testing set, despite the shortest runtime. This is due to adjusting the object's pose in the air being prone to dropping, unlike GtL, which has the ground as support and only needs to execute a static grasping. Compared with these two alternatives, our approach flexibly obtains an excellent balance between efficiency and efficacy, outperforming the baseline ‘Ours ($\chi_1=1$)’ by a large margin in success rate and surpassing the performance of the baseline ‘Ours ($\chi_1=0$)’ significantly in terms of task running cost.

\paragraph{Guidance of reward design.}
We adopt both grasp reward $r_\text{grasp}$ and efficiency reward $r_\text{effici}$ to inform the policy of grasping preferences. The grasping reward $r_\text{grasp}$  is a combination of generalized $Q_1$, success signal $G$, and execution time $r_{\text{time}}$ while we adopt $r_{\text{len}}$ and $r_{\text{step}}$ serving as efficiency reward to encourage grasping without unnecessary pin movements. We conduct an ablation study by separately removing these reward terms to investigate the impact of each component.  The conducted experiments include: ‘w/o $G$’ eliminates the success signal $G$; ‘w/o $Q_1$’ removes the generalized $Q_1$ used for measuring grasp stability;  ‘w/o $r_{\text{time}}$’ omits $r_{\text{time}}$,  designed for encouraging a decrease in execution time; ‘w/o $r_{\text{len}}$’ involves the removal of $r_{\text{len}}$, designed for encouraging shortening the pin movements, and ‘w/o $r_{\text{step}}$’ eliminates  $r_{\text{step}}$, designed for encouraging few grasp steps. We can observe that disabling each single reward component adversely affects the performance of the grasping policy, both in the training and testing sets. The policy trained without $G$ attains the lowest success rate among individual reward component ablations,  emphasizing its pivotal role in guiding successful grasps. This importance arises from the primary goal of preventing the object from falling during pin movements. The results of ‘w/o $Q_1$’ reflect that the removal of the $Q_1$ reward decreases the grasp stability compared with ‘Ours’. Disabling the execution time reward $r_{\text{time}}$ incurs the highest task running cost. Meanwhile, rewards $r_{\text{len}}$  and $r_{\text{step}}$, designed to minimize pin movements, also contribute to the quick completion of the task.

\paragraph{Necessity of state representation.}
We underscore the significance of the proposed state representation, which integrates object information, gripper information, and gripper-object interaction, by separately training policies with disabling corresponding components. Specifically, ‘w/o $o$’ disables the object information  $o$, ‘w/o $g$’ disables the gripper information $g$, and ‘w/o $f$’ disables the state representation of the gripper-object interaction $f$. We also provide results with only the state representation of the gripper-object interaction $f$, denoted as ‘only $f$’. 
The removal of either $g$ or $f$ reduces the success rate while inclining the policy towards utilizing only the more time-consuming GtL method to accomplish tasks. The underlying rationale is that both $g$ and $f$ contain information about the gripper itself, removing either of them would diminish the policy's ability to assess the current situation. This makes it more challenging to execute adaptive adjustments in the air.

To further validate the rationality of the detailed state representations mentioned in Section~\ref{sec:state_action_representation}, we conduct three additional ablation versions, including ‘w/o $world$’ denoted as the elimination of utilizing the global coordinate system to represent the gripper information $g$ and gripper-object interaction information $f$; ‘w/o $one\ hot$’ as the removal of one-hot indicator $f_c$ in $f$, and ‘w/o $pin\ pos$’ omitting the pin positions in the world frame $f_p$ and the object frame $f_p^o$. The decreases in the success rate of these experiments validate the importance of providing sufficient information for grasping execution. Additionally, we conduct a comparison to an alternative state representation using RGB-D images. From the results, one can observe that when using only RGB-D images as the full state, our method achieves a success rate of $73.19\%$ on the testing set and $71.75\%$ on the training set. This suggests that our method performs well even with this single-view partial observation, though not as well as the designed representation of state. We will elaborate on how our state representation can be applied in physical practice in Section~\ref{sec:5_4}.

\paragraph{Significance of curriculum learning.}
One of our key contributions is the introduction of a curriculum learning technique directing the policy model to gain proficiency in both GtL and GwL skills. To highlight the significance of our curriculum learning approach, we conduct specific ablations on different stages of the curriculum. We stipulate that ‘w/o $CL$’ denotes training without the two-stage curriculum; ‘w/o $pre$’ represents two-stage training without the pre-collected GwL experiences, ‘w/o $CL\  \& \ pre$’ removes both the two-stage curriculum and the pre-collected experiences, and ‘Reward $CL$’ represents a curriculum framework based on reward shaping, where all grasping rewards except execution time are initially applied, and the time penalty is later introduced. As shown in Table~\ref{tab:ablation_table}, we can see that removing the two-stage curriculum impairs the flexibility of the policy to switch grasping modes. The policy achieves a successful grasp only through the more challenging GwL motions, which significantly reduce the grasp success rate.
Eliminating the pre-collected GwL experiences leads to a relatively lower success rate on the testing set, given the lack of prior experience in air adjustments to guide policy learning. The same reasons apply to ‘w/o $CL\  \& \ pre$’ which learns policies without guidance about both ground adjustments and air adjustments. ‘Reward $CL$’ also yields a lower success rate than our curriculum design, highlighting that our two-stage curriculum is better suited for facilitating learning across GtL and GwL skills.

\subsection{Comparison Results.}
We design an adaptive geometry-aware pin-pression gripper that can adjust its finger shapes according to the variations of the geometry and pose of the target object. To demonstrate the necessity of learned grasping policies and the effectiveness of our gripper, we make comparisons with other alternative grippers and the passive grasping approach. 
\label{sec:5_3}

\paragraph{Comparison with alternative grippers.}
\begin{table}
\centering
\caption{Success rate comparisons on artificially generated CAD and scanned object datasets. All shapes are unseen for our trained policies.}
\label{tab:compare_others}
\setlength{\tabcolsep}{0.75mm}{
\begin{tabular}{c|c|c|c|c|c|c} 
\hline
\multirow{2}{*}{Method} & \multicolumn{3}{c|}{CAD models} & \multicolumn{3}{c}{Scanned models}  \\ 
\cline{2-7}
                        & ShapeNet& Thingi10K & ABC            & YCB    & BigBIRD & KIT         \\ 
\hline\hline
WSG50                   & $43.28\%$   & $41.34\%$   & $40.28\%$         & $58.22\%$ & $47.20\%$  & $41.09\%$           \\ 
\hline
Fit2form                & $78.82\%$   & $76.96\%$   & $68.12\%$         & $78.21\%$ & $84.00\%$  & $80.62\%$    
\\ 
\hline
Random            & $22.37\%$   & $28.86\%$   & $23.34\%$         & $35.44\%$ & $31.20\%$  & $24.03\%$           \\ 
\hline
Passive            & $60.36\%$   & $67.51\%$   & $56.94\%$         & $60.76\%$ & $68.80\%$  & $55.81\%$          \\ 
\hline
Ours                     & $\textbf{81.54\%}$   & $\textbf{80.42\%}$   & $\textbf{76.91\%}$         & $\textbf{81.01\%}$ & $\textbf{88.00\%}$  & $\textbf{87.60\%}$           \\
\hline
\end{tabular}}
\end{table}

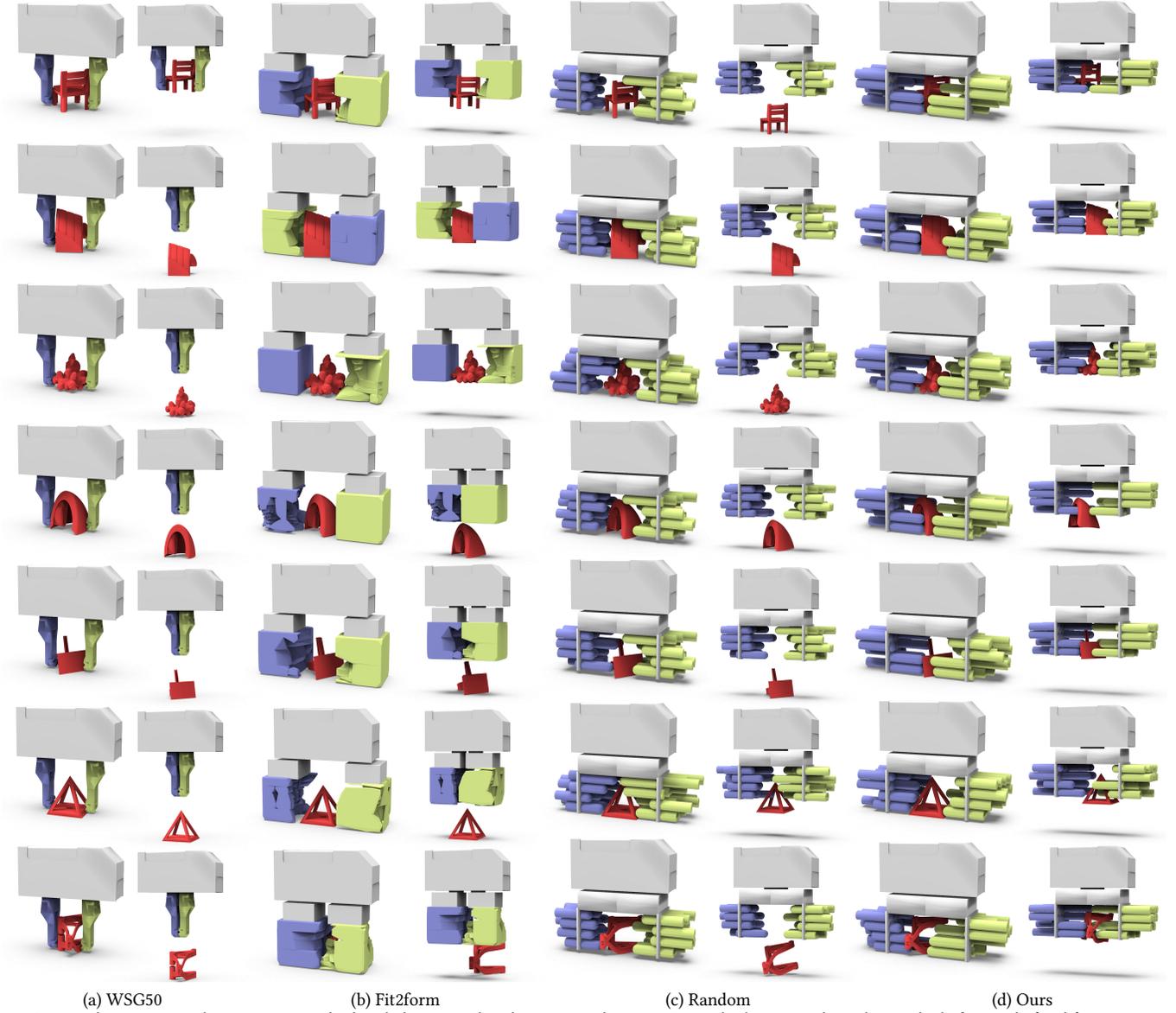
\begin{figure*}[!t]
  \centering
  \begin{overpic}[width=1.0\linewidth,tics=10]{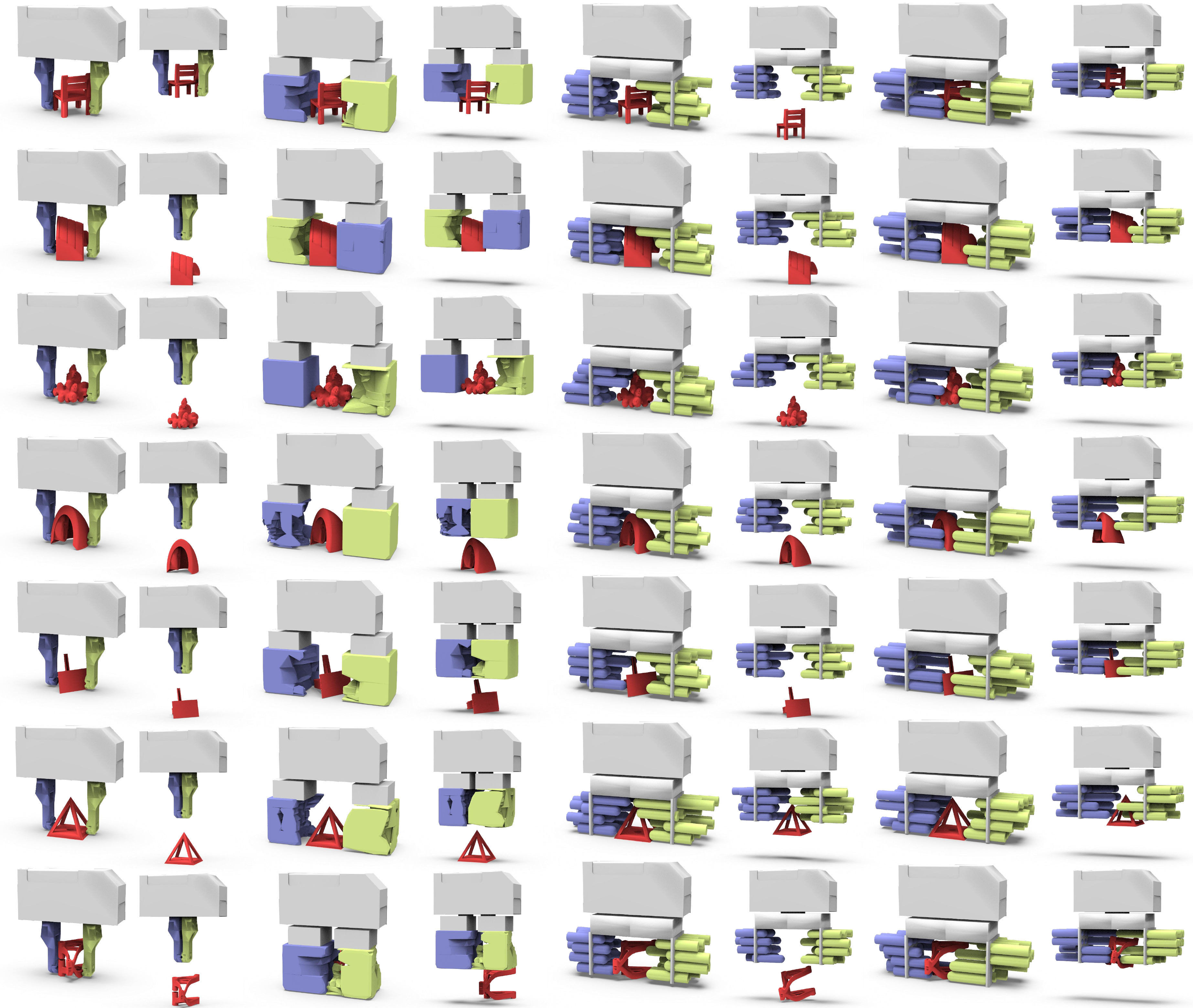}
  \put(7,-2){\small (a) WSG50}
  \put(30,-2){\small (b) Fit2form}
  \put(57,-2){\small (c) Random}
  \put(85,-2){\small (d) Ours}
  \end{overpic}
  \caption{Visual comparisons between our method and alternative baselines on six datasets. For each object, we show the results before and after lifting it. Note that our method is capable of generating successful grasps for both artificially generated CAD models and real-world scanned models.}
   \label{fig:Comparison_examples}
\end{figure*} 
For general gripper grasping, previous methods often utilize either universal parallel-jaw grippers for general-purpose manipulation, designed without consideration of the target geometry, or task-specific grippers, which are geometry-aware and execute grasping by closing all gripper fingers. To facilitate a more comprehensive comparison, we categorize the baseline variants based on these two gripper design approaches and evaluate their success rates separately. For the general-purpose grippers, we compare our method with the WSG50 gripper~\cite{wsg50}, which features specialized components on its fingertips to enhance friction and increase the contact area for stable grasping. For the task-specific grippers, we compare our method with a 3D generative design approach, Fit2form~\cite{fit2form}, which generates pairs of finger shapes to execute grasps for fixed object poses. The experiment setting for baselines closely follows the implementation from Fit2Form and performs the top-down grasp. We also compare our method with multi-finger robot grippers, such as the two-finger gripper (EZGripper), the three-finger gripper (Barrett hand), and the dexterous hand (Shadow hand) (see Table 2 and Section C in supplemental material). Additionally, to further demonstrate the effectiveness of our learned policies, we compare our method with two other approaches: the random grasping policy, which randomly executes pin movements throughout the entire task process, and the passive grasping method, which adjusts finger shapes passively by directly applying external force.

We evaluate the grasp success rate of each approach on both Computer-Aided Design (CAD) and real-world scanning. For artificially generated CAD models, we test on the ShapeNet dataset, the Thingi10K dataset, and the ABC dataset. For the scanned shapes, we test on the YCB dataset, the BigBIRD dataset, and the KIT dataset. Specifically, we randomly select 1511 shapes from ShapeNet, 1542 shapes from Thingi10K, and 1728 shapes from ABC, as well as using all available shapes from the YCB dataset, the BigBIRD dataset, and the KIT dataset. Note that all shapes are unseen for the policies that only learn on the training set from Section~\ref{sec:5_2}.

Table~\ref{tab:compare_others} reports the quantitative comparisons to those alternative baselines on six different datasets. We also provide visual comparisons between our method and these baselines in Figure~\ref{fig:Comparison_examples}. Obviously, our method consistently outperforms all the baselines on all the datasets. We observe that these alternative grippers underperform against variations in object geometry or poses. In contrast, our method is capable of adaptively modifying its finger shape, guided by the interaction between the gripper and the object. It can also be observed that the random grasping policy may produce sparse and insufficient contacts between objects and the gripper fingers, failing to achieve perfect force closure for the target objects. Additionally, our method also achieves good performance across different grasping directions in terms of success rate (See Figure 3 in supplemental material). Overall, these comparative results further underscore the advantages of our geometry-aware pin-pression gripper design.
\begin{table*}[!t]
\caption{
Comparison with the WSG50 gripper and the Passive Grasp method of directly extending all pins. Here, Chal-H represents the category of flat objects, while Chal-T represents objects with inclined surfaces or tetrahedron-like shapes.
}
\vspace{-10pt}
\label{tab:compare_passive}
\setlength{\tabcolsep}{1.2mm}{
\begin{tabular}{l|cc|cc|cccc|cccc}
\hline
& \multicolumn{2}{c|}{Train}
& \multicolumn{2}{c|}{Test}
& \multicolumn{4}{c|}{Chal-H}
& \multicolumn{4}{c}{Chal-T}
\\ \hline
Method         
& \multicolumn{1}{c|}{S $(\%)\uparrow$}       
& \multicolumn{1}{c|}{$Q1 \uparrow$}       
    
& \multicolumn{1}{c|}{S $(\%)\uparrow$}     
& \multicolumn{1}{c|}{$Q1 \uparrow$}      

& \multicolumn{1}{c|}{S $(\%)\uparrow$}     
& \multicolumn{1}{c|}{$Q1 \uparrow$}      
& \multicolumn{1}{c|}{GwL $(\%)$} 
& GtL $(\%)$ 
& \multicolumn{1}{c|}{S $(\%)\uparrow$}     
& \multicolumn{1}{c|}{$Q1 \uparrow$}      
& \multicolumn{1}{c|}{GwL $(\%)$} 
& GtL $(\%)$ 
\\ 
\hline
\hline
WSG50  
& \multicolumn{1}{c|}{39.24} 
& \multicolumn{1}{c|}{0.2181} 

& \multicolumn{1}{c|}{36.08} 
& \multicolumn{1}{c|}{0.2226} 

& \multicolumn{1}{c|}{26.00} 
& \multicolumn{1}{c|}{0.2012}
& \multicolumn{1}{c|}{0.00} 
& 100.00
& \multicolumn{1}{c|}{22.42} 
& \multicolumn{1}{c|}{0.2217} 
& \multicolumn{1}{c|}{0.00} 
& 100.00
\\ 
\hline
Passive Grasp      
& \multicolumn{1}{c|}{58.97}   
& \multicolumn{1}{c|}{0.3414}  
    
& \multicolumn{1}{c|}{60.82} 
& \multicolumn{1}{c|}{\textbf{0.3560}} 

& \multicolumn{1}{c|}{36.00} 
& \multicolumn{1}{c|}{0.3146} 
& \multicolumn{1}{c|}{0.00}     
& 100.00   
& \multicolumn{1}{c|}{44.82} 
& \multicolumn{1}{c|}{0.3280} 
& \multicolumn{1}{c|}{0.00}     
& 100.00   
\\ 
\hline
Active Grasp (Ours) 
& \multicolumn{1}{c|}{\textbf{82.96}}   
& \multicolumn{1}{c|}{\textbf{0.3506}}  

& \multicolumn{1}{c|}{\textbf{82.47}} 
& \multicolumn{1}{c|}{0.3456} 
 
& \multicolumn{1}{c|}{\textbf{72.00}} 
& \multicolumn{1}{c|}{\textbf{0.3264}} 
& \multicolumn{1}{c|}{28.00} 
& 72.00 
& \multicolumn{1}{c|}{\textbf{63.79}} 
& \multicolumn{1}{c|}{\textbf{0.3526}} 
& \multicolumn{1}{c|}{75.87} 
& 24.13 
\\ 
\hline
\end{tabular}
}
\end{table*}

\begin{figure*}[!t]
  \centering
  \begin{overpic}[width=1.0\linewidth,tics=10]{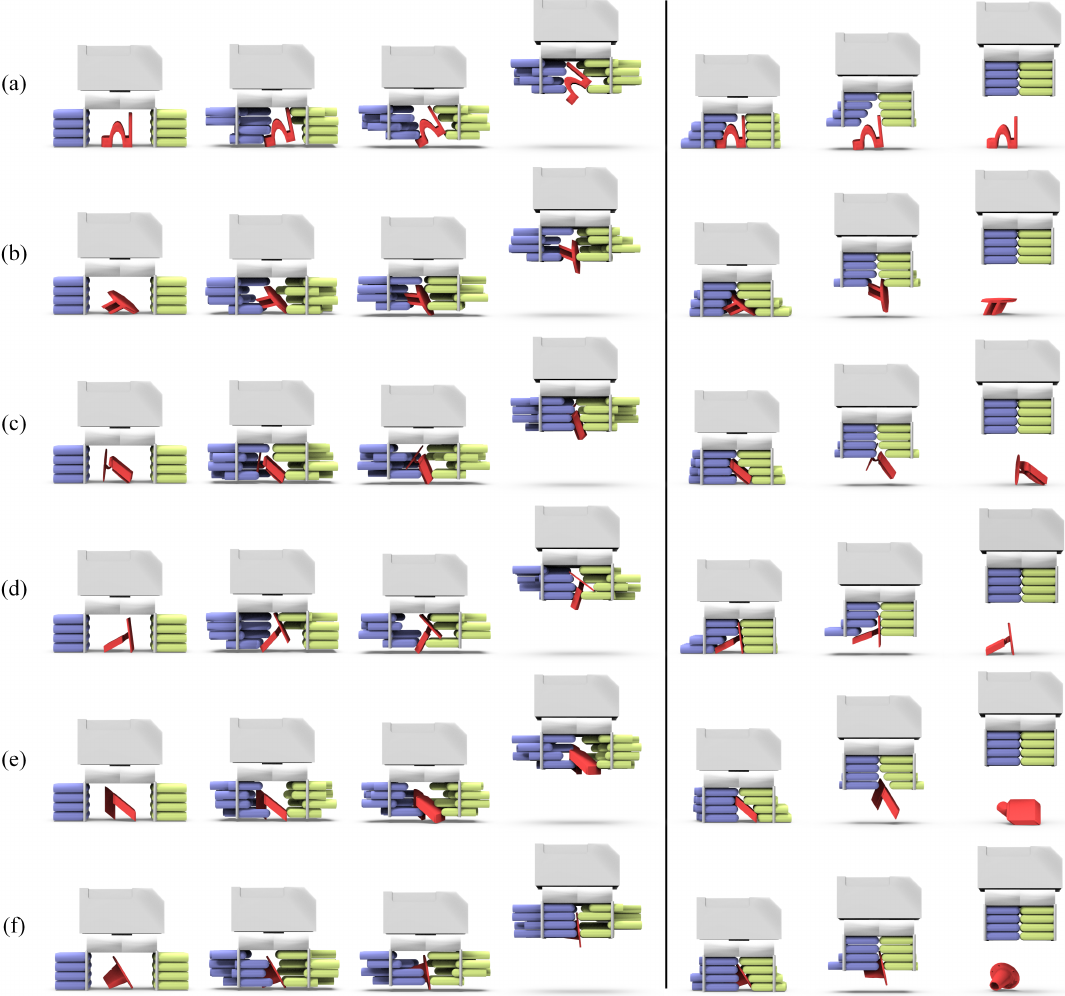}
  \put(31,-2){\small Ours}
  \put(79,-2){\small Passive Grasp}
  \end{overpic}
  \caption{
  Visualization of our approach (left) and the passive grasping method with all pins extended  (right). We display the initial configurations of the gripper and several key frames of the grasping processes. 
  For the Passive Grasp method, we showcase the moment of object and gripper detachment.
  }
   \label{fig:compare_all_closed}
   \vspace{-7pt}
\end{figure*}

\paragraph{Comparison with passive grasping method.}
Previous efforts turn to directly introducing external force 
to passively adjust finger shapes~\cite{robotics2019versaball, scott1985omnigripper}. 
These gripper fingers typically operate in a fixed mode, facing challenges for shape variations. For example, flat objects usually offer limited contact areas for the gripper, especially when lying on the ground. Objects with inclined surfaces or tetrahedral shapes also present difficulties due to inadequate force closure. Inspired by these, we construct a challenging dataset to further demonstrate the necessity of our online grasping approach, i.e., grasping objects with adaptive gripper adjustment. The principle for constructing the dataset is that the objects are either flat or have inclined surfaces. We finally obtain a Chal-H dataset with $50$ flat shapes and a Chal-T dataset consisting of objects with inclined surfaces or tetrahedron-like shapes. All shapes are unseen for the grasping policies during the training. More details of the challenge dataset can be found in Section B of supplemental material. We compare our learned policies with directly extending all pins to passively conform to the shape of the target object. The quantitative comparisons are reported in Table~\ref{tab:compare_passive}, and the qualitative comparisons are illustrated in Figure~\ref{fig:compare_all_closed}.

Table~\ref{tab:compare_passive} reveals that our learned policies exhibit superior adaptability in comparison to the conventional parallel jaw gripper (WSG50) and passive grasping. Our gripper achieves $72.00\%$ grasp success rate on flat objects and $63.79\%$ on tetrahedron-like shapes, compared to $36.00\%$ and $44.82\%$ for Passive Grasp, respectively. WSG50 grippers also achieve low success rates on both datasets, i.e., $26.00\%$ for Chal-H and $22.42\%$ for Chal-T, demonstrating the necessity of our adaptive gripper design.

As illustrated in Figure~\ref{fig:compare_all_closed}, the passive grasping approach struggles due to the unidirectional extension of all pins, which often fails to establish effective force closure, causing the object to be pushed out of the gripper and dropped. It can be observed from Figure~\ref{fig:compare_all_closed} that our method is capable of grasping the target objects through online adjustments. Throughout the GwL phase, the object's pose evolves through in-hand rotation, and the grasp configuration undergoes refinement for better force closure through dynamic adjustments. This also underscores the advantages of active grasping.

Additionally, we uncover an intriguing phenomenon by observing the proportion between GwL and GtL motion. For objects featuring flat shapes, the GtL skill is utilized more frequently (constituting $72.00\%$ of cases), thereby facilitating the precise identification of optimal contact points. In contrast, for objects characterized by inclined surfaces or tetrahedron-like shapes, our algorithm tends to the utilization of GwL motion (accounting for $75.87\%$ of instances)  with in-air adjustments, to alter the relative pose between the object and the gripper, thereby ensuring more effective grasping. This highlights the capability of our approach in shape perception and adaptability. 

\begin{figure}[!t]
  \centering
  \begin{overpic}[width=1
  \columnwidth,tics=10]{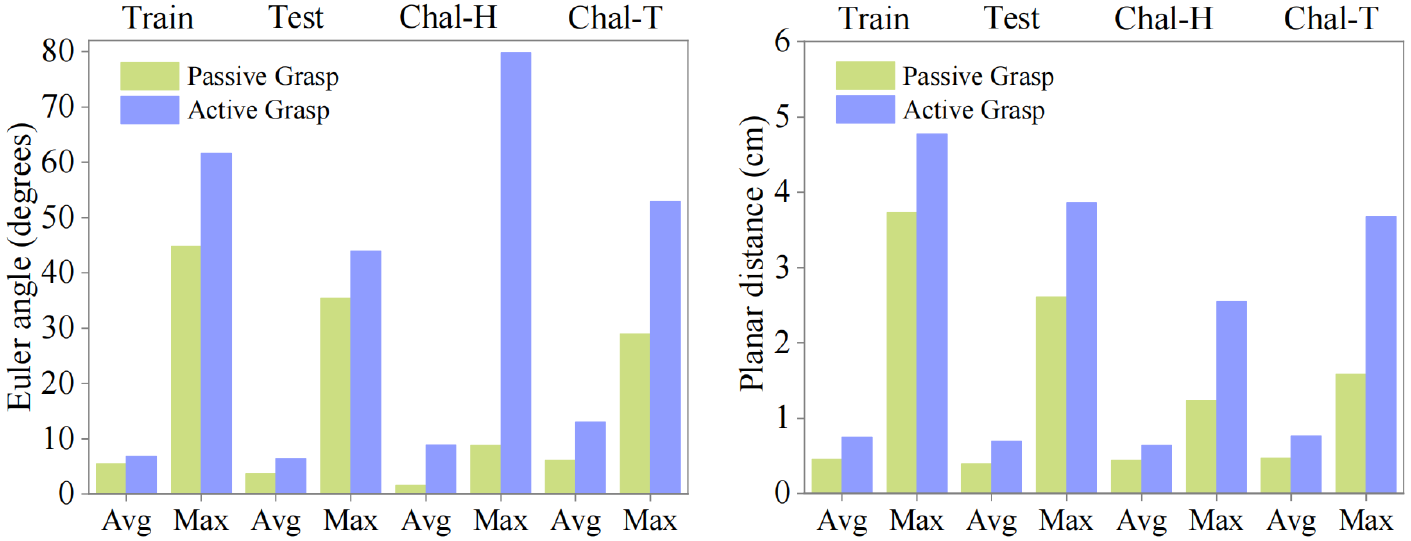}
  \put(22,-3){\small (a)}
  \put(75,-3){\small (b)}
  \end{overpic}
  \caption{
  Flexibility analysis on Active Grasp (Ours) and Passive Grasp methods in successfully grasped objects. (a) statistics the Euler angle variation (measured in degrees); (b) statistics positional variation along the xoy-plane (planar distance measured in centimeters). For each dataset, we statistic both the mean (Avg) and maximum (Max) values. 
  }
   \label{fig:compare_flexible}
   \vspace{-7pt}
\end{figure} 
\subsection{Analysis of Pin-Pression Gripper Design}
\label{sec:5_4}
\paragraph{Gripper flexibility analysis.}
To illustrate the flexibility of our pin-pression gripper, we measure the variations of  Euler angles and planar distances of the successfully grasped objects during the grasping procedure. We compare our method with the Passive Grasp method and we measure these pose changes before and after grasping. We calculate the Euclidean distance in the horizontal xoy plane and measure the planar shifts in centimeters. For the Euler angle, we convert the quaternions of the objects into rotation matrices, and then compute the angular difference between the two rotation matrices. As shown in Figure~\ref{fig:compare_flexible}, our method exhibits richer pose variation than passive grasping, proving that our RL-based policy enhances grasping performance through in-hand adjustments. 

\begin{figure}[!t]
  \centering
  \begin{overpic}[width=1\columnwidth,tics=10]{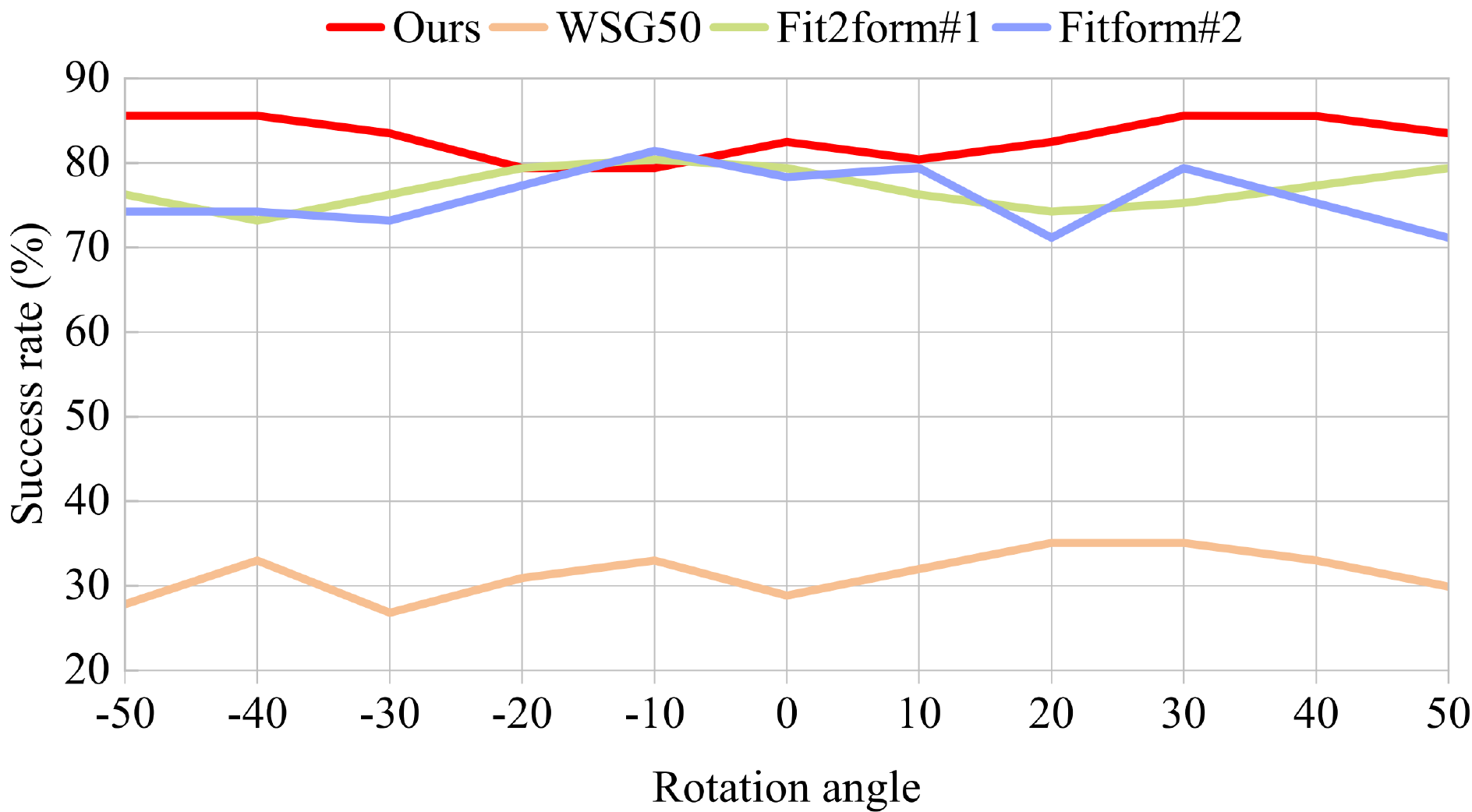}
  \end{overpic}
  \caption{
  Comparisons of the robustness to object rotations. The test objects undergo various rotations around the z-axis.
  } \label{fig:robustness_to_object_orientation}
  \vspace{-7pt}
\end{figure} 
\paragraph{Robustness to object rotation.}
To further demonstrate the robustness of our approach to object rotation, we compare its performance against the WSG50 gripper and Fit2form method, measuring success rate at different levels of object rotation. Specifically, all objects are rotated by $\pm 10^{\circ}$, $\pm 20^{\circ}$, $\pm 30^{\circ}$, $\pm  40^{\circ}$, and $\pm 50^{\circ}$ around the z-axis, respectively, and then the corresponding success rates are calculated. We make sufficient comparisons with the Fit2form method. For each object, we measure the grasping success rate by utilizing different gripper shapes specifically generated for the corresponding rotation (Fit2form\#1). We also employ a consistent gripper shape for grasping an object with its various rotations (Fit2form\#2). The success rates of these methods are reflected in Figure~\ref{fig:robustness_to_object_orientation}.

It can be observed that our adaptive pin-pression gripper is nearly always superior in handling object rotations, even if a significant object rotation is applied.
The general-purpose WSG50 gripper cannot fit well to dynamic variations of object rotations due to the fixed finger shape. This results in consistently exhibiting the worst performance under perturbations of different rotations. We also note that the Fit2form method is sensitive to the rotation of objects, even for Fit2form\#1 where the gripper shapes are specifically generated for the corresponding rotation. More visualization results about the robustness of our pin-pression gripper to arbitrary object poses can be found in Section D of supplemental material.

\begin{figure}[!t]
  \centering
  \begin{overpic}[width=1.0\columnwidth,tics=10]{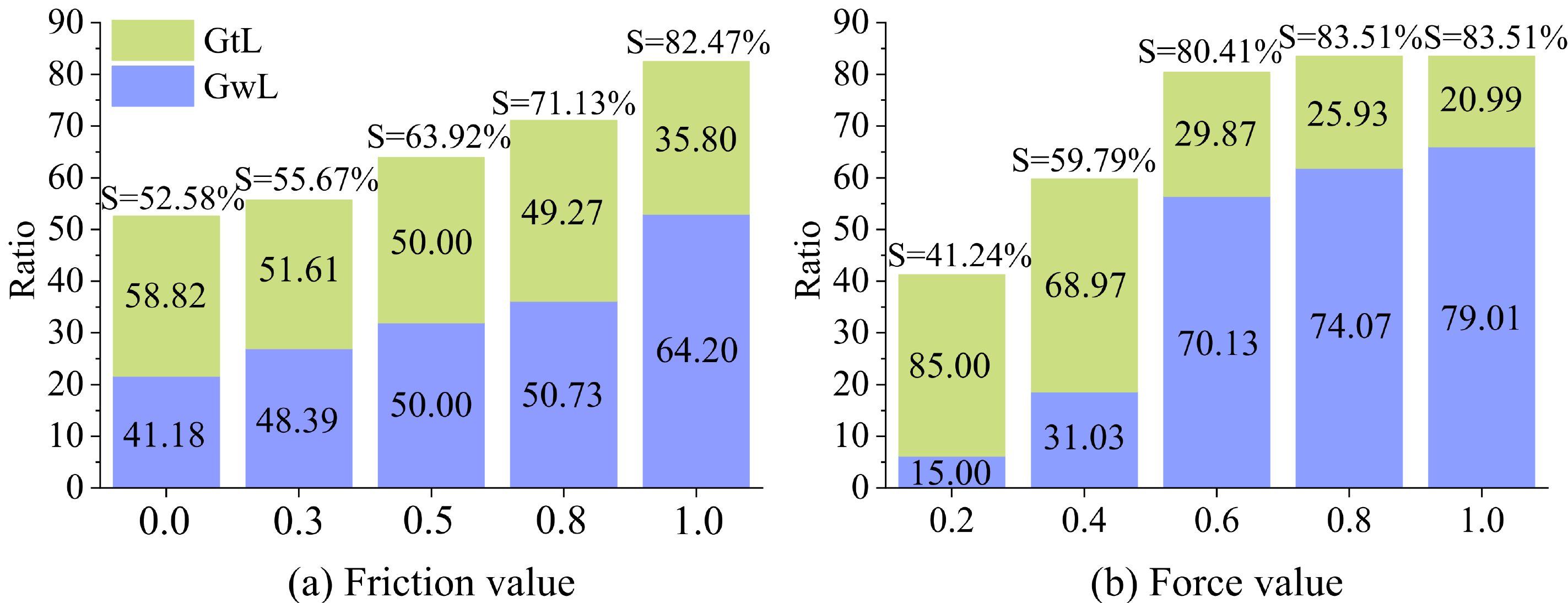}
  \end{overpic}
  \caption{The influences of gripper friction and grasp force on the performance of our method. We evaluate the grasp success rate and the percentages of both GtL and GwL motions, across different dynamic parameters.}
  \label{fig:Analysis_of_different_grasp_force_friction}
  \vspace{-10pt}
\end{figure} 
\paragraph{Friction and grasp force analysis.}
We also analyze the impacts of gripper friction and grasp force on the performance of our learned policies, aiming to unveil interesting aspects of grasping behavior. 
We evaluate our method on the grasp success rate and the percentages of GtL and GwL motions.  
In the friction analysis,  we vary the gripper friction from 0.0 to 1.0 and retrain the policy network accordingly. The results with different gripper friction parameters are illustrated in Figure~\ref{fig:Analysis_of_different_grasp_force_friction}a. As anticipated, the success rate consistently improves as we increase the friction parameter, suggesting that gripper friction plays a crucial role in achieving a successful grasp. Moreover, an interesting observation in grasping mode is noteworthy: with an increase in friction, there is a notable rise in the percentage of GwL motion. This suggests that our method retains the capability to flexibly adopt appropriate grasping modes, coordinating pin motion with the target object to successfully grasp it.
For the grasp force analysis, we evaluate a total of five grasp forces. The related results of our method under different grasp forces are shown in Figure~\ref{fig:Analysis_of_different_grasp_force_friction}b. It can be seen that grasp force is also a pivotal factor in the performance of our method for adaptively manipulating the gripper toward a successful grasp. We also evaluate the robustness of our method to variations in object mass and friction. These experiments are conducted to assess the success rate of our method on the testing data (see Figure 5 in supplemental material).

\paragraph{Exploration of pin density.}
\begin{table}[t]
\centering
\caption{
Performance of our pin-pression gripper with various pin densities.
}
\label{tab:differ_grid}
\setlength{\tabcolsep}{3.2mm}{
\begin{tabularx}{0.4\textwidth}{c|ccc}
\hline
Pin density 
& \multicolumn{1}{c|}{$S (\%) \uparrow$} 
& \multicolumn{1}{c|}{$Q1 \uparrow$}     
& $T (s) \downarrow$              
\\ 
\hline
\hline
$3\times3$    
& \multicolumn{1}{c|}{50.51} 
& \multicolumn{1}{c|}{0.3416} 
& 594.24 
\\ 
\hline
$4\times4$    
& \multicolumn{1}{c|}{82.47} 
& \multicolumn{1}{c|}{0.3456} 
& 634.38   
\\ 
\hline
$5\times5$    
& \multicolumn{1}{c|}{88.66} 
& \multicolumn{1}{c|}{0.3458} 
& 688.67
\\ 
\hline
\end{tabularx}
}
\end{table}

We explore pin arrays of different densities, including $3\times3$, $4\times4$, and $5\times5$. We maintain the overall area of the gripper and only redistribute the pins for the different densities. We retrain the $3\times3$ and $5\times5$ gripper configurations and report the grasp success rate and the sum running time cost in Table~\ref{tab:differ_grid}. In our pursuit to design a pin-pression gripper that performs well across a wide variety of shapes while minimizing the number of pin actuators, we found that a 4×4 grid size is sufficiently flexible to achieve a general grasping effect, with an average success rate over $82.00\%$ and significantly outperform other density alternatives.

\paragraph{Sim-to-Sim transfer.}
\begin{table}[t]
\centering
\caption{Sim-to-Sim transfer experiment. We evaluate the robustness of our learning-based approach by transferring it to a different physics simulator Isaac Gym~\cite{Makoviychuk2021IsaacGH}.} 
\label{tab:sim2sim}
\begin{tabular}{c|c|c|c} 
\hline
\multirow{2}{*}{Method} & \multirow{2}{*}{$S$ (\%) $\uparrow$} & \multicolumn{2}{c}{Time cost (s)  $\downarrow$} \\
\cline{3-4}
                        &                          & Computation & Simulation  \\ 
\hline\hline
WSG50                   & $47.42\%$                    & $0.00$           & $765.81$             \\ 
\hline
Fit2form                & $70.10\%$                    & $477.56$            & $806.95$             \\ 
\hline
Ours                     & $73.19\%$                    & $0.79$           & $784.47$             \\
\hline
\end{tabular}
\end{table}

Transferring grasping capabilities to simulation environments with different physics is of utmost importance for validating the generality of our approach, encompassing both the hardware design of our pin-pression gripper and our learning-based grasping algorithm. Different simulators may exhibit variations in physical dynamics, such as subtle differences in collision detection, object modeling, and numerical integration methods. To show the robustness of our method, we compare it with other alternative methods by directly transferring them to a different physics simulator. Specifically, we replicate the gripper grasping environment from PyBullet~\cite{coumans2019}, employing the open-source Bullet Physics engine, to a new test scene in Isaac Gym~\cite{Makoviychuk2021IsaacGH} with the PhysX engine~\cite{nvidia_physx}. We evaluate both the grasp success rate and the sum running time cost for all test instances in Isaac Gym, and we report the results in Table~\ref{tab:sim2sim}. Our geometry-aware approach sustains considerable grasping performance in the new simulation environment, surpassing other methods in terms of success rate. Furthermore, our method requires less computational and simulation time compared to the Fit2form method, which is known for its time-consuming nature for customized gripper shape generation.

\paragraph{Sim-to-Real transfer.}
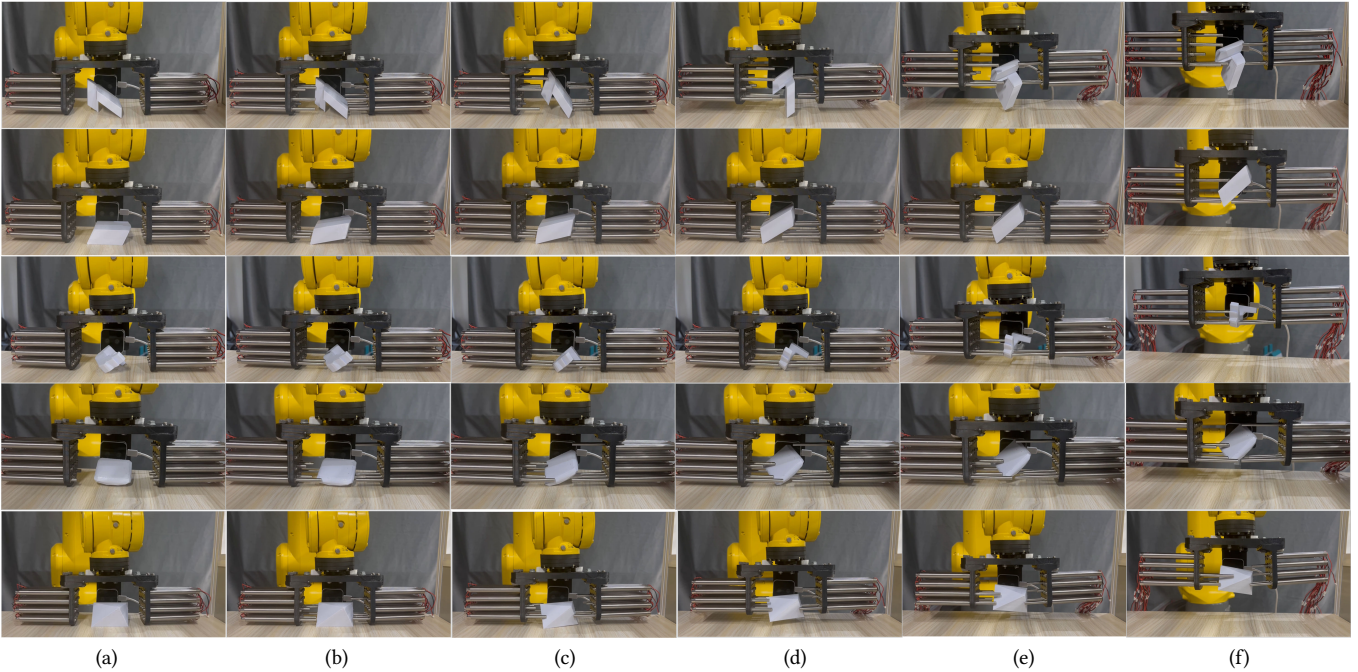
\begin{figure*}[!t]
  \centering
  \begin{overpic}[width=1.0\linewidth,tics=10]{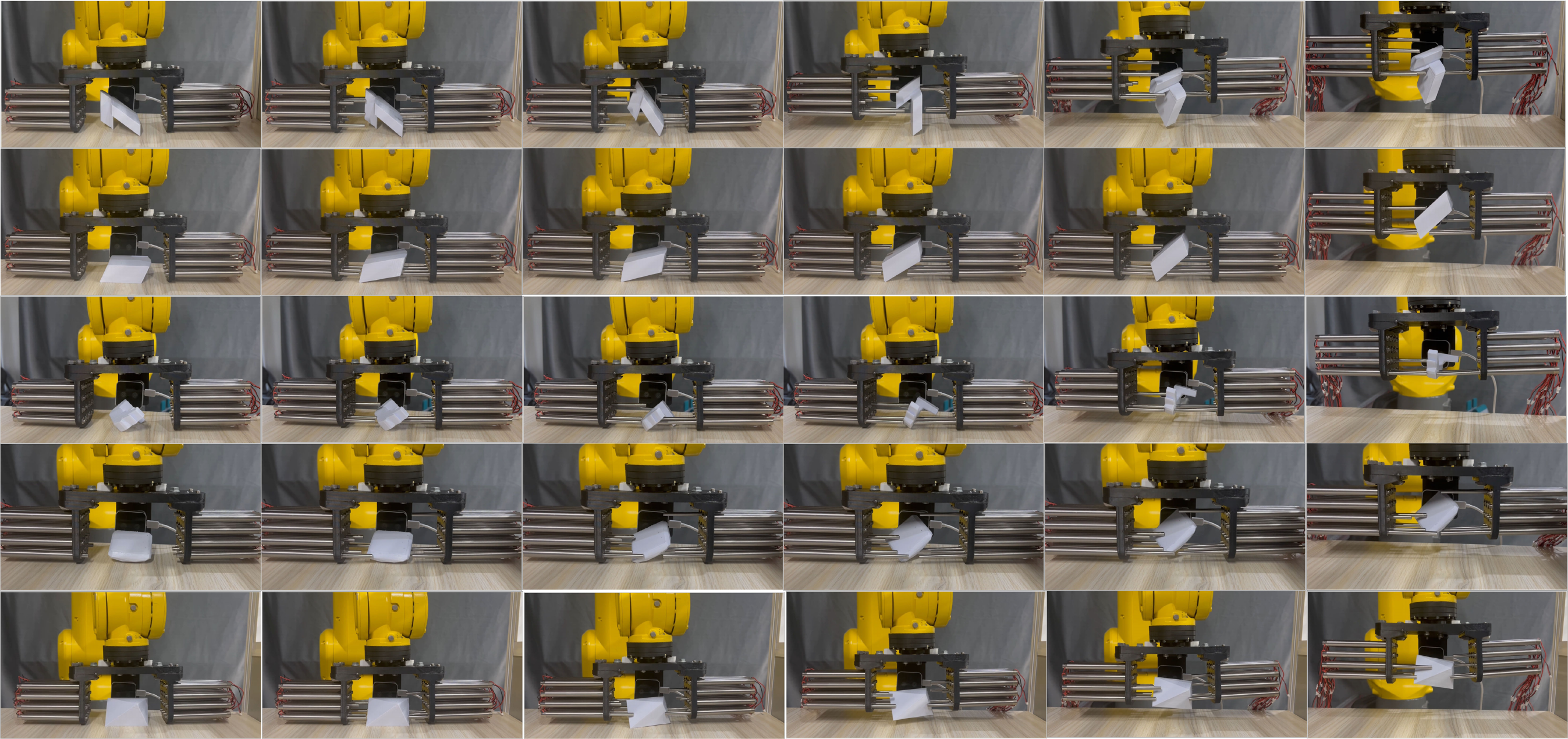}
  \put(7,-2){\small (a) }
  \put(24,-2){\small (b) }
  \put(41,-2){\small (c) }
  \put(58,-2){\small (d) }
  \put(75,-2){\small (e) }
  \put(91,-2){\small (f) }
  \end{overpic}
  \caption{
  Physical experiment of sim-to-real transfer. From (a) to (f) show several key frames of the dynamic grasping processes of our pin-pression gripper on five object models. The gripper approaches the target object from the top, forms a basic closure against the object with pin movements, and lifts the object while further performing in-hand re-orientation to improve the grasping stability.}
  
   \label{fig:sim_to_real}
   \vspace{-5pt}
\end{figure*} 
We further validate the performance of our approach on a real robotic platform. Transferring grasping capabilities to real-world environments presents significantly greater challenges for both gripper manufacturing and grasping policy learning compared to simulations. On the one hand, we must ensure that the physically manufactured gripper aligns with our design, particularly for the automatic and accurate control of the pin's extension and retraction. On the other hand, we need to ensure that the grasping policy trained in simulation can be effectively transferred to the real world. 

To achieve this, we use an electric actuator as the pin of the gripper finger, which allows us to automatically and accurately control and read the pin's movement distance (\textit{pin reading}). We then physically fabricate a pin-pression gripper featuring $4\times 4$ pin arrays, ensuring it is proportionally consistent with our gripper design and exact in size. This gripper is also equipped with an in-hand RGB-D camera for capturing the object shape. To adapt our grasping policy to the real-world environment, we employ a two-stage teacher-student training paradigm~\cite{Chen2022VisualDI}. First, we train a teacher policy network using reinforcement learning with full state information captured in simulation. Subsequently, we train a student policy network via DAGGER algorithm~\cite{ross2011reduction}, which takes virtual RGB-D observations and pin readings as inputs and is optimized to match the actions predicted by the teacher policy network. Lastly, we use real-world observations, including pin readings from each actuator and RGB-D images from the camera, as inputs to the student policy network, thereby transferring the grasping capability to the real world. 

The physical experiment of sim-to-real transfer is illustrated in Figure~\ref{fig:sim_to_real}. We showcase several key steps of dynamically adjusting the extension and retraction of pins throughout the grasping process, applied to five object models. As illustrated, target objects are successfully grasped and lifted by the dynamic grasping skill of our physically manufactured pin-pression gripper. This confirms the effectiveness and practicality of our approach in real-world applications. For more details on the experiment implementation, hardware specifications, and the results of the physical experiment, please refer to Section E in supplemental material.

\section{DISCUSSION AND FUTURE WORK}\label{sec:discussion}
We concurrently investigate gripper design and dynamic grasping skills for 3D objects featuring diverse geometry and poses. We have developed an innovative pin-pression gripper capable of adaptively adjusting its shape to grasp objects with various geometries and topologies and under arbitrary poses. We also propose an RL-based method to learn adaptive grasping policies along with a curriculum learning technique to facilitate online grasping using different grasping modes. Through extensive quantitative and qualitative experiments on diverse novel object datasets, our method demonstrates the capability to handle complex shapes and achieve dynamic robot grasping with a high success rate and efficiency. We also physically manufacture the pin-pression gripper based on the design and highlight sim-to-real results on a real robot platform, confirming that our method can be effectively transferred to real-world applications.
\begin{figure}[!t]
  \centering
  \begin{overpic}[width=1\columnwidth,tics=10]{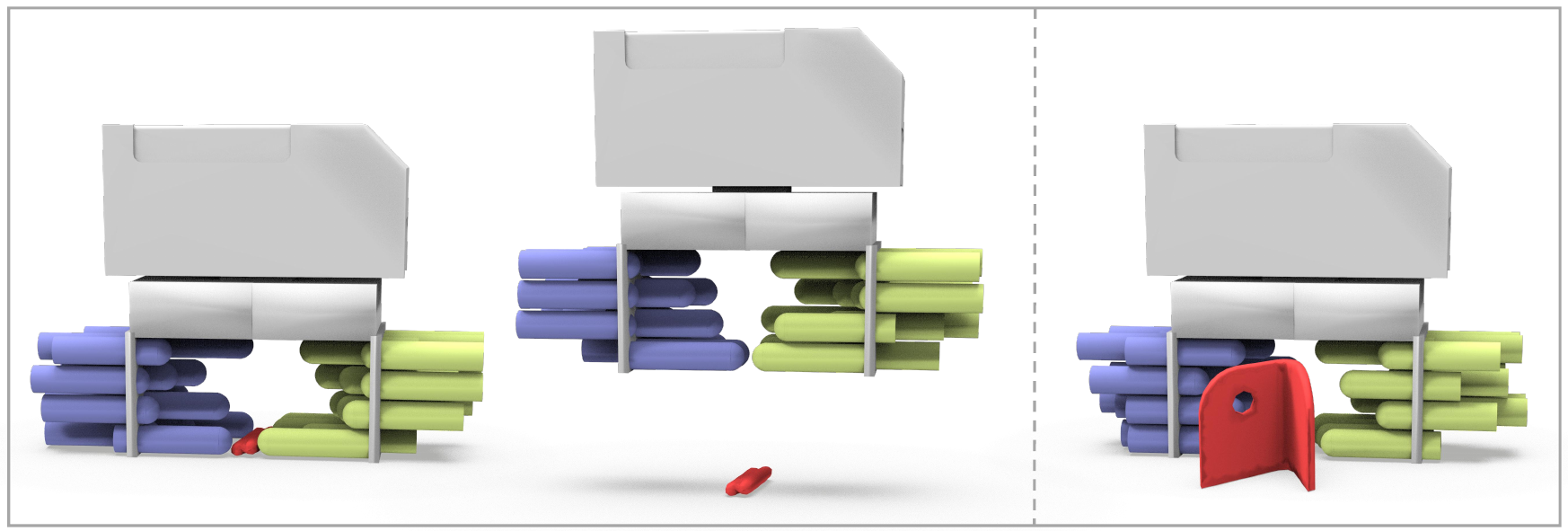}
  \put(32,-3){\small (a)}
  \put(82,-3){\small (b)}
  \end{overpic}
  \caption{Failure cases. Our gripper is not adept at handling certain extremely thin shapes (a), due to the cylindrical shape formulation of the pins and the limited pin grain. The same reasons may also lead to objects with curved surfaces (b) being pushed out.
  }
   \label{fig:limitation}
   \vspace{-18pt}
\end{figure} 

As an initial endeavor to develop a novel pin-pression gripper to tackle the significant challenges of adaptive grasping, our work still has certain limitations. 
Firstly, to achieve adaptive grasping, the pins on our gripper necessitate a broader range of adjustments, leading to the scales of target objects being limited. Secondly, our gripper may encounter difficulties with extreme cases, such as thin objects like chopsticks (refer to Figure~\ref{fig:limitation}a), owing to the constrained space available for the gripper to perform the adjustments. We also observe that certain objects with curved surfaces might be pushed out as a result of the force generated along the x-axis by the interaction of pins with their surfaces (see Figure~\ref{fig:limitation}b).

Additionally, our grasping process is a multi-objective optimization that simultaneously considers task success, grasp stability, and efficiency. Since we prioritize success and stability, the efficiency reward is given a lower weight, allowing some redundant actions. Further improving the execution efficiency of the pin-pression gripper is an important area for future work. In parallel, we are engaged in applying our gripper to real-world applications in daily life scenarios~\cite{li2024llm} and industrial automation~\cite{zhao2023learning}.
\begin{acks}
This work was supported in parts by NSFC (62325211, 62495081, 62272082, 12494554, 62132021), the Joint Fund General Project of Liaoning Provincial Department of Science and Technology (2024-MSLH-352), and the Major Program of Xiangjiang Laboratory (No. 23XJ01009).
\end{acks}

\bibliographystyle{ACM-Reference-Format}
\bibliography{sample-bibliography}

\end{document}